\documentclass[10pt,twocolumn,letterpaper]{article}
\usepackage[pagenumbers]{cvpr} 

\usepackage{bm}
\usepackage{multirow}
\usepackage{float}
\usepackage[ruled,vlined]{algorithm2e} 

\definecolor{tableblue}{RGB}{162, 185, 235}
\definecolor{tablegreen}{RGB}{6, 214, 160}
\definecolor{softred}{RGB}{255, 112, 112}
\definecolor{softgreen}{RGB}{112, 200, 112}

\definecolor{cvprblue}{rgb}{0.21,0.49,0.74}
\usepackage[pagebackref,breaklinks,colorlinks,allcolors=cvprblue]{hyperref}

\title{FreeCond: Free Lunch in the Input Conditions of Text-Guided Inpainting}

\author{Teng-Fang Hsiao, Bo-Kai Ruan, Sung-Lin Tsai, Yi-Lun Wu, Hong-Han Shuai\\
National Yang Ming Chiao Tung University\\
{\tt\small tfhsiao.ee13@nycu.edu.tw, bkruan.ee11@@nycu.edu.tw, tsai412504004.ee12@nycu.edu.tw}\\
{\tt\small yilun.ee08@nycu.edu.tw, hhshuai@nycu.edu.tw}}

\begin{document}
\twocolumn[{%
\renewcommand\twocolumn[1][]{#1}%
\maketitle
\begin{center}
    \centering
    \captionsetup{type=figure}
    \includegraphics[width=\linewidth]{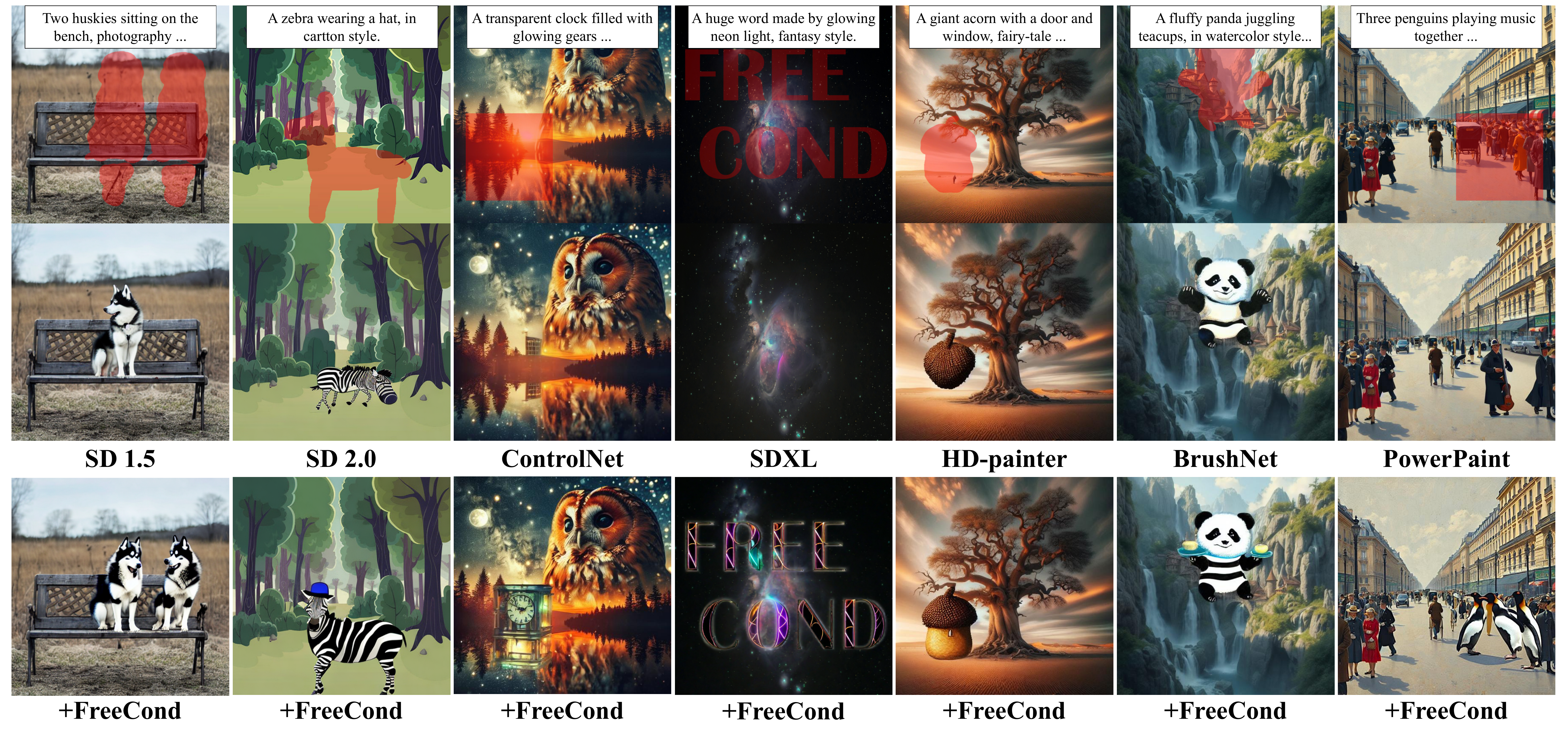}
     \caption{Comparison of T2I inpainting methods with FreeCond, applied across various mask types: ``multi-masks'' (column 1 and 4), ``precise masks'' (columns 2, 5, and 6), and ``rough masks'' (columns 3 and 7) with complex prompts and unrelated image contexts. By integrating FreeCond, existing inpainting baselines obtain better ``instruction-following'' performance.}
\label{fig:teaser}
\end{center}%
}]

\begin{abstract}
In this study, we aim to determine and solve the deficiency of Stable Diffusion Inpainting (SDI) in following the instruction of both prompt and mask. Due to the training bias from masking, the inpainting quality is hindered when the prompt instruction and image condition are not related. Therefore, we conduct a detailed analysis of the internal representations learned by SDI, focusing on how the mask input influences the cross-attention layer. We observe that adapting text key tokens toward the input mask enables the model to selectively paint within the given area. Leveraging these insights, we propose FreeCond, which adjusts only the input mask condition and image condition. By increasing the latent mask value and modifying the frequency of image condition, we align the cross-attention features with the model’s training bias to improve generation quality without additional computation, particularly when user inputs are complicated and deviate from the training setup. Extensive experiments demonstrate that FreeCond can enhance any SDI-based model, e.g., yielding up to a 60\% and 58\% improvement of SDI and SDXLI in the CLIP score. The code and appendix are available in our repository at \url{https://github.com/basiclab/FreeCond} \footnote{Due to arXiv file size limitations, we provide an abbreviated version of the paper here; the full version can be accessed in the repository.}.
\end{abstract}
\section{Introduction}
\label{sec:introduction}

\begin{figure}
    \centering
    
    \begin{subfigure}[b]{0.48\linewidth}
        \centering
        \text{\footnotesize\textit{``\textcolor{cyan}{A fluffy panda} \textcolor{Orchid}{juggling teacups}}''}
        \includegraphics[width=\linewidth]{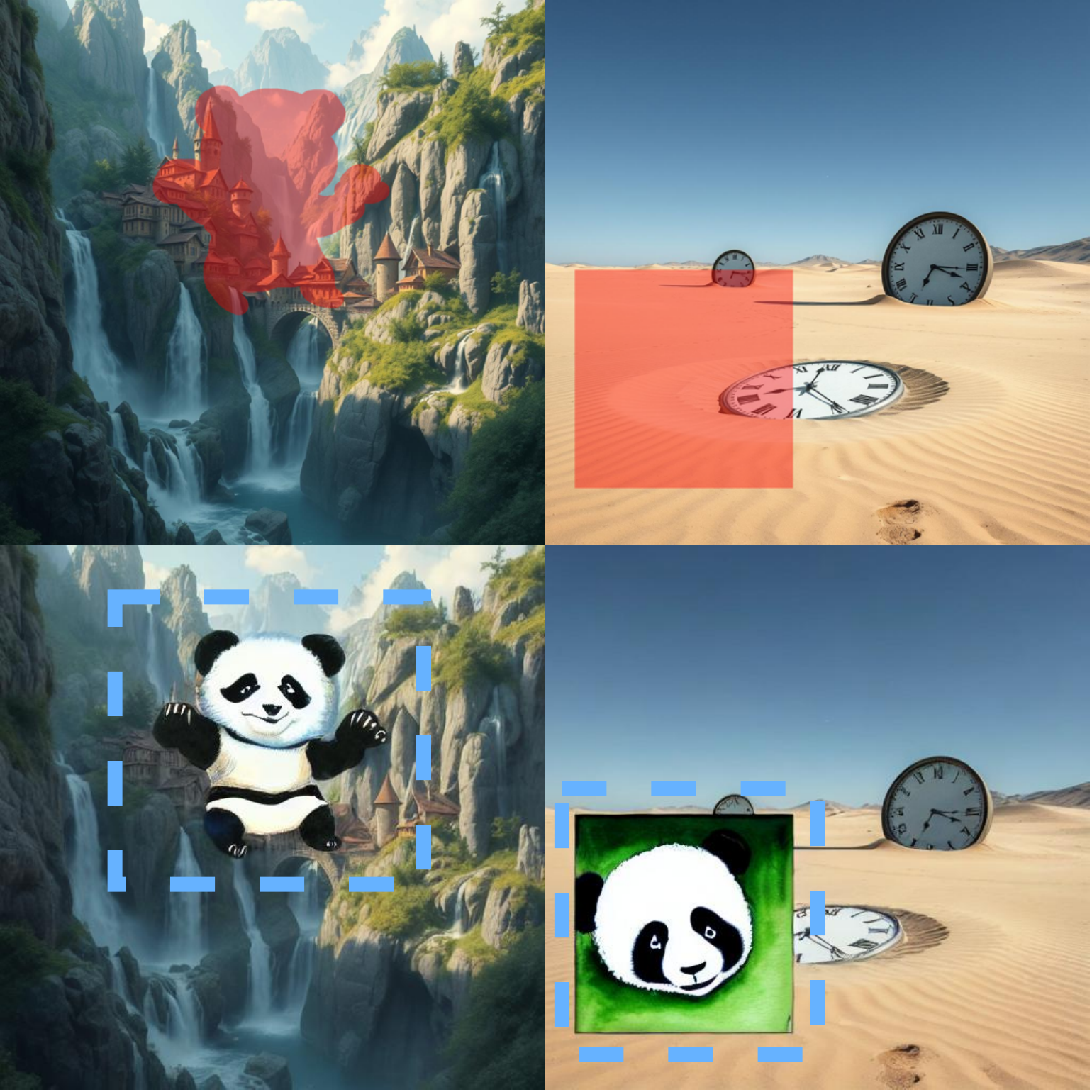}
        \caption{BrushNet}
    \end{subfigure}
    \begin{subfigure}[b]{0.48\linewidth}
        \centering
        \text{\footnotesize\textit{``\textcolor{cyan}{Three penguins} \textcolor{Orchid}{playing music}}...''}
        \includegraphics[width=\linewidth]{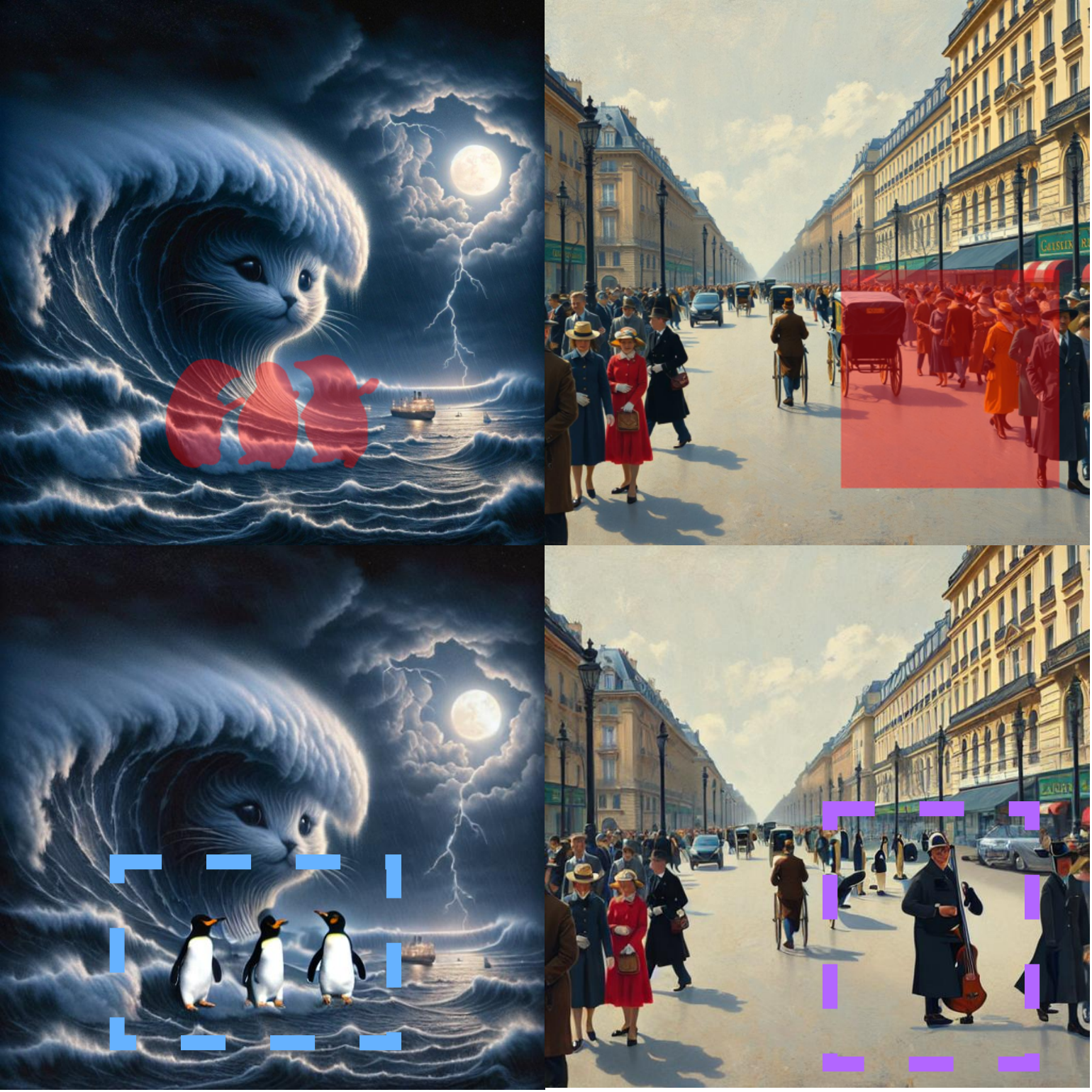}
        \caption{PowerPaint}
    \end{subfigure}
    \caption{Comparison of existing SOTA methods. BrushNet rigidly follows the mask instructions but only partially adheres to the prompt. PowerPaint produces outputs that are harmonious with the image context but at the cost of reduced prompt-adherence. FreeCond addresses these limitations, as shown in \cref{fig:teaser}.}
    \label{fig:bp_output}
    \vspace{-10pt}
\end{figure}

Text-to-image (T2I) inpainting seeks to fill specified masked areas based on user-provided text prompts. Stable Diffusion Inpainting (SDI), a tailored adaptation of Stable Diffusion~\cite{stablediffusion}, is widely used for its effectiveness in achieving high-quality inpainting aligned with text prompts. However, the SDI training process employs a random masking strategy, which often hinders the model's ability to follow prompts accurately and fit masks precisely, especially when the prompt lacks contextual relevance to the surrounding image. We refer to these dual issues of “prompt-adherence” and “mask-fitting” collectively as the “instruction-following” problem: the model prioritizes contextual coherence over generating content that strictly adheres to both the complex prompt details and the user-specified mask, as illustrated in the second row of \cref{fig:teaser}.

To address these limitations, methods such as HD-Painter~\cite{hdpainter}, BrushNet~\cite{brushnet}, and PowerPaint~\cite{powerpaint} have been developed. BrushNet, for example, leverages segmentation-based training data and a ControlNet-like structure, allowing it to learn a direct link between the input mask and prompt. PowerPaint incorporates training with dilated segmentation masks and task-specific tokens, enabling flexible object inpainting that better conforms to varied shapes. While these approaches effectively reduce the “mask-fitting” issue, they primarily optimize for simple prompts and often lack the generalization capability required for complex prompt adherence, as shown in \cref{fig:bp_output}. Observing the limitations of training-only modifications, \textbf{we propose to directly modify the model's behavior by adjusting its learned mechanism.}

In this paper, we contend that effective instruction-following relies on the differential noise predictions: \textit{conditional versus unconditional}, modulated via classifier-free guidance (CFG)~\cite{cfg}. This differential is notably manifested in the outputs of the cross-attention layer, where prompt tokens receive significantly higher attention within the masked area than in surrounding regions. Consequently, the query and key features within the cross-attention layer must dynamically adapt to the input mask. This adaptation focuses on generating coherent new content within the masked regions while concurrently preserving the integrity of the surrounding context by selectively enhancing features related to the mask in the corresponding channels.

To address these challenges, we propose FreeCond, a training-free method that requires no extra computation. Specifically, FreeCond filters the high-frequency components of the image condition in the early diffusion steps, reducing the contextual information of the image condition. Furthermore, FreeCond induces a stronger feature shift in the cross-attention layer by scaling the mask condition, enabling stronger activation of the masked area. By adjusting the input conditions, FreeCond significantly enhances both prompt-adherence and mask-fitting while preserving overall harmony. Notably, FreeCond, as a more general form of noise prediction function, can be seamlessly integrated with other SDI-based methods~\cite{hdpainter, powerpaint, brushnet, controlnet, sdxl}, as it directly enhances the original SDI backbone


Finally, we propose FCIBench (FreeCond Inpainting Benchmark), a new benchmark with 600 inpainting pairs, to handle complex inpainting scenarios. Compared with existing inpainting benchmarks~\cite{coco,brushnet}, FCIBench includes precise masks, rough masks, and multi-masks, along with complex prompts that are unrelated to image condition, as shown in the first row of \cref{fig:teaser}. This variety enables a more comprehensive evaluation of SDI across diverse inpainting conditions. Our expanded benchmark thus helps a thorough assessment of the performance of different models across varied prompts and mask configurations. Experimental results demonstrate that FreeCond consistently improves performance across models and benchmarks, achieving a 60\% increase in CLIP score~\cite{clip} of SDI backbone and a 1\% increase of existing SOTA. The contributions can be summarized as follows.

\begin{itemize}
    \item We conduct an in-depth analysis of the SDI model’s mechanism, enhancing the interpretability by examining its learned bias of relying on image context and its capability to selectively inpaint within the masked area.
    \item We introduce FreeCond, a novel noise prediction function, that addresses the instruction-following limitations of SDI-based models without adding computational overhead, especially in scenarios where the complex prompt instruction is unrelated to the image condition.
    \item We provide FCIBench to evaluate inpainting methods in scenarios across precise mask, rough mask, and multi-mask settings, along with complex prompts that are unrelated to image conditions, extending the evaluation to more diverse scenarios.
\end{itemize}
\section{Related Works}

\subsection{Image Inpainting}
Image inpainting focuses on repainting specified regions while ensuring coherence with the surrounding image. Various non-text-guided inpainting methods have been developed to achieve this~\cite{contextual_attention, free_form, aggregated_contextual, inpainting_vqvae, pluralistic, repaint, in-n-out, maskgit}, alongside the emergence of text-to-image inpainting methods~\cite{bld, controlnet, zero_shot, sdedit, uni_paint, edit_bench}. SDI~\cite{stablediffusion} pioneered the integration of a random masking strategy into its training objective, producing harmonized outputs. However, this approach often prioritizes image conditioning over adherence to instructions. To improve this, recent methods~\cite{smartbrush, powerpaint, brushnet, hdpainter} have introduced solutions focused primarily on enhancing mask-fitting. Despite these advancements, these methods often lack prompt -adherence when handling complex instructions. In contrast, FreeCond leverages insights into the inpainting mechanism to improve instruction-following across any SDI-based inpainting model, achieving a balanced performance between mask-fitting and prompt-adherence.

\subsection{Unveiling the Mechanism of T2I Models}
T2I diffusion models possess powerful image-generation capabilities. To fully harness this potential, recent works have explored various training-free modifications based on in-depth analyses of different components~\cite{masactrl,style_injection,zstar,tfgph,prompt-to-prompt, a-cat-is-a-cat, understanding_mechanism,compositional,edit_friendly,understanding_t2i,freeenhance}. Notably, FreeU~\cite{freeu} reveals that the skip-connection primarily retains texture details, while the backbone captures more semantic information. By adjusting the balance between them, FreeU enhances semantic accuracy with minimal impact on detail and computational costs.

Our analysis reveals that the image and mask conditions play similar roles: the image condition provides contextual details, while the mask condition regulates prompt influence. By modulating both conditions, we achieve improved instruction-following without additional costs, providing a “free lunch” in performance enhancement.
\section{Analysis of SDI model}
\label{sec:model_analysis}
In this section, we provide an in-depth analysis of the SDI model. Firstly, we identify the training bias of the SDI model in \cref{sec:d1} that the model heavily relies on the image condition to generate the content. This mechanism can cause the generated output unrelated to the input prompt. Secondly, we demonstrate in \cref{sec:d2} that increasing the size of the mask provides SDI with more potential solutions to integrating prompt instructions into the image context, resulting in outputs that more closely follow the instruction. Finally, in \cref{sec:d3}, we test the hypotheses that interpret the internal mechanics of the SDI model, with a particular focus on the relationship between the inpainting conditioning and the cross-attention layer. The analysis of the SDI model helps us understand the factors leading to successful instruction-following inpainting.


\subsection{Stable Diffusion Inpainting Model (SDI)}
\label{sec:pre}
The SDI model receives a prompt $p$, an image $I \in \mathbb{R}^{H \times W \times 3}$, and a mask $M \in \mathbb{R}^{H \times W}$ that specifies the inpainting area. To prevent the model from copying content directly from $I$, the masked area of $I$ is set to zero, yielding $I^c = (\bm{1}-M) \odot I$. Since the UNet operates in VAE~\cite{vae} latent space, $I^c$ is encoded as the image condition $z^c = \mathcal{E}(I^c) \in \mathbb{R}^{(H/4) \times (W/4) \times 4}$ and $\mathcal{E}$ denotes the pretrained VAE encoder. To match the input size, the SDI model uses an interpolated mask condition $M^c \in \mathbb{R}^{(H/4) \times (W/4)}$, created by downsampling $M$ with nearest-neighbor interpolation. The final inpainting noise prediction from the diffusion model is $\epsilon_{{\theta}}(z_t,z^c,M^c,t,p)$, where $\epsilon_{{\theta}}$ represents the SDI UNet model, $z_t$ is the noise latent at timestep $t \in [0,T]$, and initial noise $z_T$ is sampled from $\mathcal{N}(\bm{0},\bm{I})$. To control the influence of prompt $p$, we follow classifier-free guidance (CFG)~\cite{cfg}, modifying the noise prediction with a scaling parameter $w\in\mathbb{R}$:
\begin{align}
    \hat{\epsilon}_\theta(&z_t, z^c, M^c, t, p) = \epsilon_\theta(z_t, z^c, M^c, t, \varnothing) \nonumber \\
    &+w\big(\epsilon_\theta(z_t, z^c, M^c, t, p) - \epsilon_\theta(z_t, z^c, M^c, t, \varnothing)\big)
    \label{eq:cfg}
\end{align}


To evaluate the inpainting results in our study, we use a set of six metrics adopted from BrushBench~\cite{brushnet}. These metrics cover three key areas: \textbf{Image Quality}, measured by Image Reward (IR)~\cite{imagereward}, HPS~\cite{hps}, and Aesthetic Score (AS)~\cite{laion5b}; \textbf{Background Preservation}, assessed using PSNR and LPIPS\cite{lpips}; and \textbf{Instruction Following}, evaluated through CLIP~\cite{clip}. Additionally, we introduce a novel Intersection-over-Union (IoU) score to specifically capture the mask-fitting quality, complementing the CLIP Score by distinguishing mask accuracy from prompt-adherence. This score is computed by the IoU between input mask and auto-labeled mask via SAM~\cite{sam}, as detailed in Appendix.



\begin{figure}[t]
    \centering
    \text{\textit{``zebra''}}
    
    \begin{subfigure}[b]{0.47\linewidth}
        \centering
        \includegraphics[width=\linewidth]{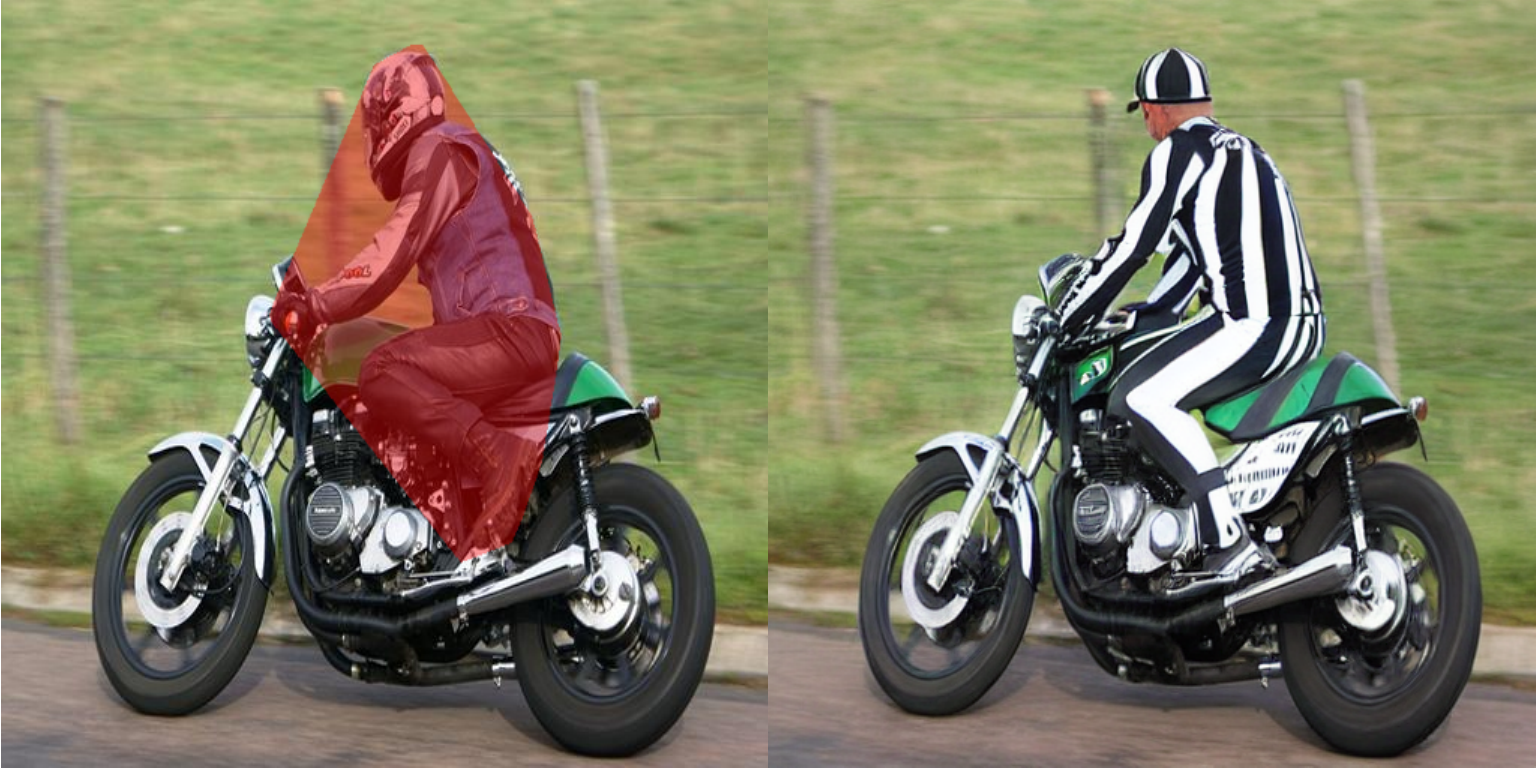}
    \end{subfigure}
    \begin{subfigure}[b]{0.47\linewidth}
        \centering
        \includegraphics[width=\linewidth]{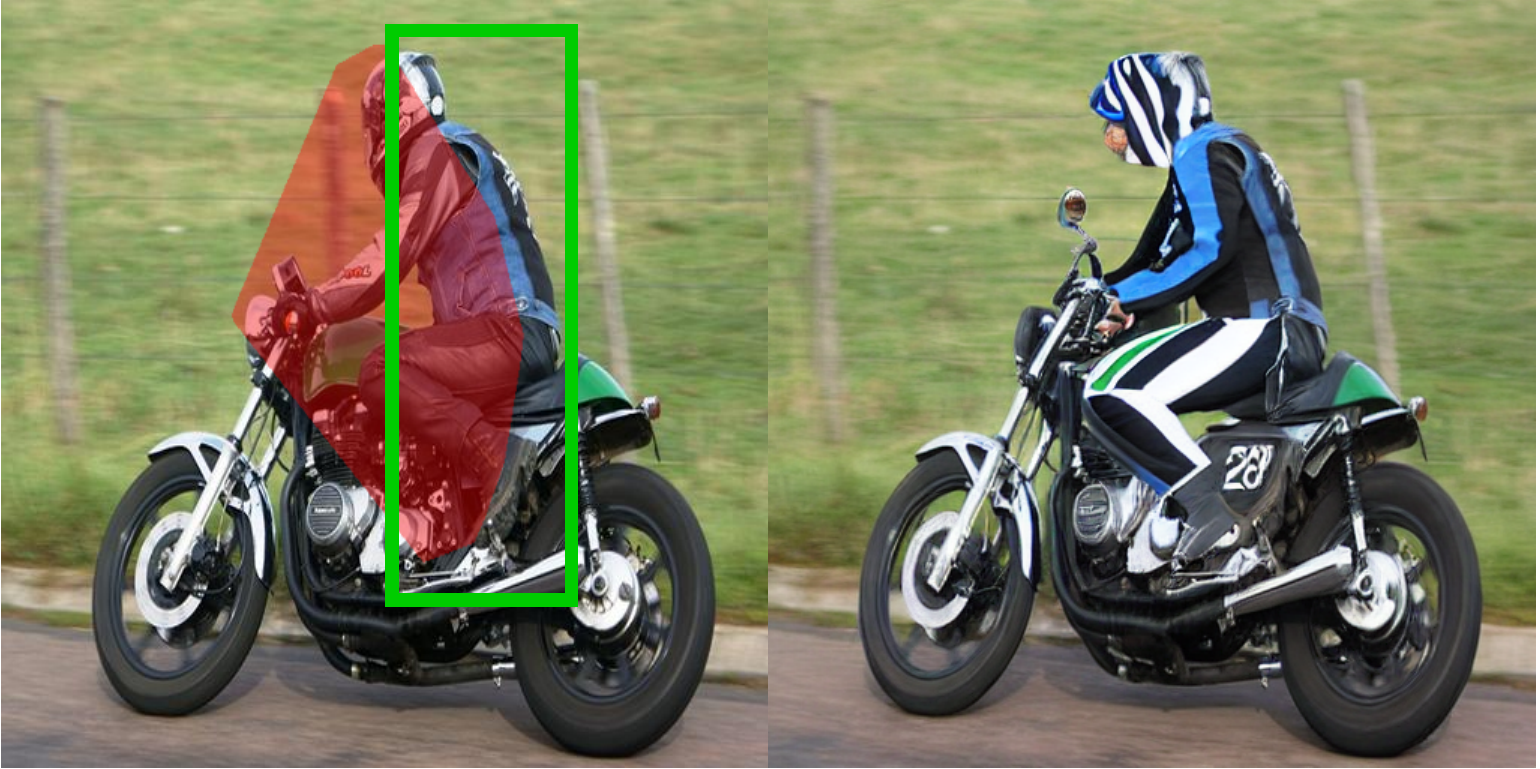}
    \end{subfigure}

    \text{\textit{``airplane''}}
    
    \begin{subfigure}[b]{0.47\linewidth}
        \centering
        \includegraphics[width=\linewidth]{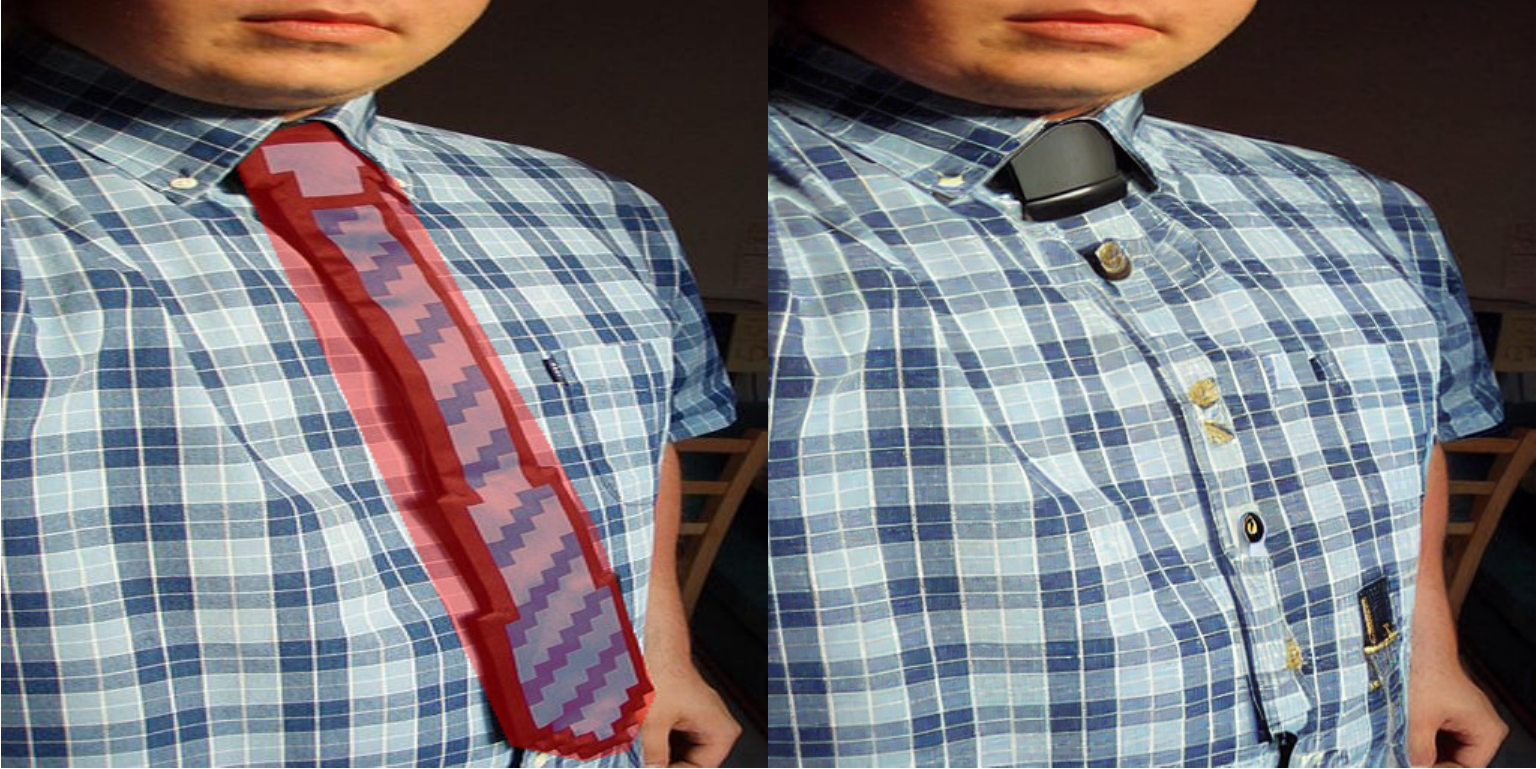}
        \caption{Random Prompt}
        \label{fig:vis_random_shift:a}
    \end{subfigure}
    \begin{subfigure}[b]{0.47\linewidth}
        \centering
        \includegraphics[width=\linewidth]{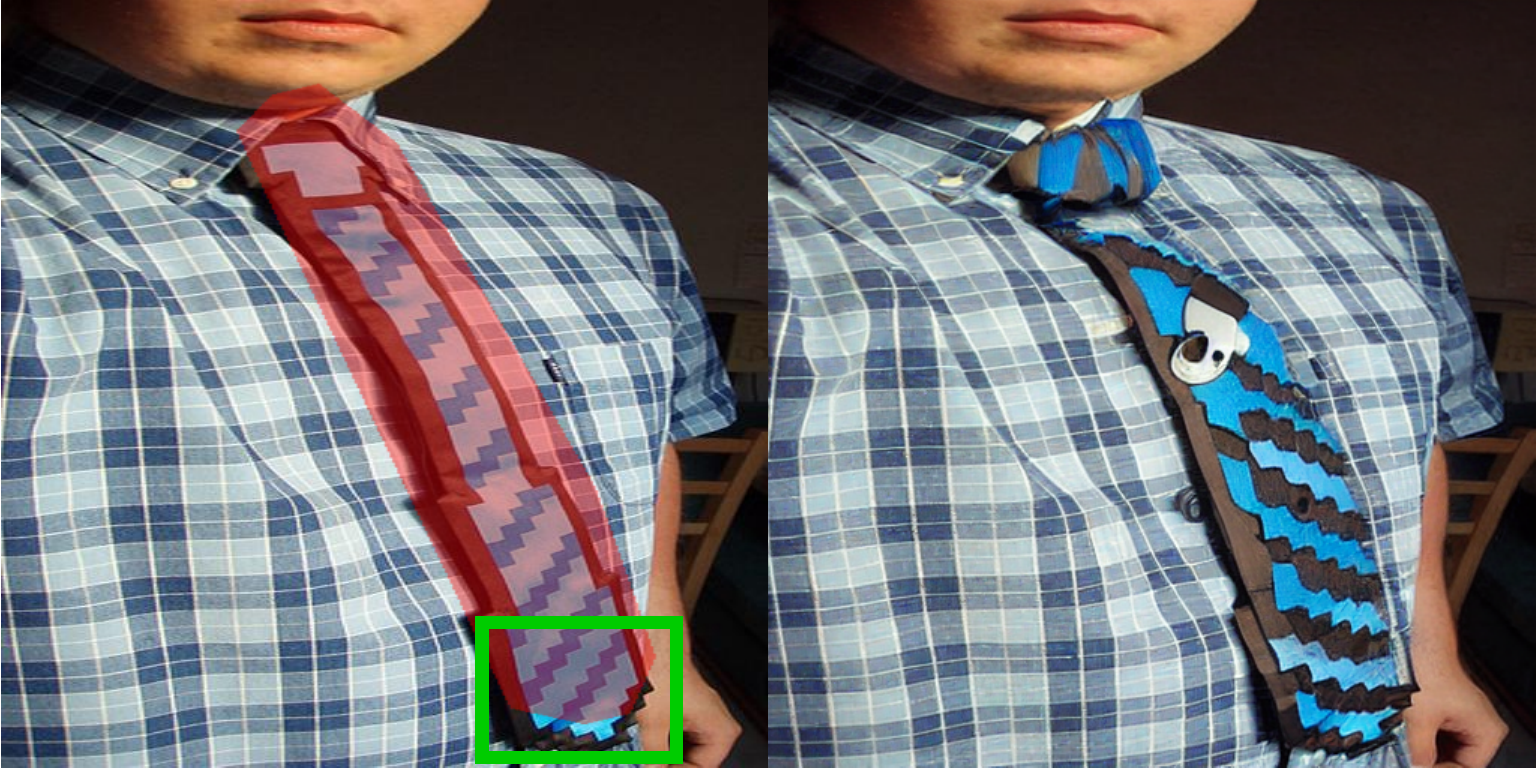}
        \caption{Random Prompt + Shift Mask}
        \label{fig:vis_random_shift:b}
    \end{subfigure}
    \caption{Visualization of contextual influence: A random prompt, unrelated to the image condition, is assigned. The input mask is shown in columns 1 and 3, along with the corresponding prompt, while shifted areas are highlighted with a \textcolor{Green}{green frame}. The resulting outputs are displayed in columns 2 and 4.}
    \label{fig:vis_random_shift}
    \vspace{-10pt}
\end{figure}

\subsection{Influence of Image Condition $z^c$}
\label{sec:d1}
The random masking strategy of SDI is via creating masked data by randomly masking 25\% of image areas in LAION-5B~\cite{laion5b}, aiming to enhance generalizability across various inputs. To investigate the mask distribution under this random strategy, we define three mask placements: “not masked,” “partially masked,” and “fully masked.” Although the exact SDI training mask distribution is not accessible, our analysis on the COCO dataset as a surrogate reveals that, with a 25\% mask coverage, over 80\% of training data falls under the “not masked” or “partially masked” categories (see the Appendix for details). Thus, we hypothesize that \textbf{the random masking design optimizes SDI for maintaing overall image harmony rather than strict prompt-adherence}. For instance, in the second row of \cref{fig:vis_random_shift:a}, when prompted with \textit{“airplane”} instead of the actual ground-truth \textit{“tie”}, SDI ignores the prompt and generates contextually consistent but unrelated content.

To validate this, we select 600 mask and ground-truth pairs from COCO~\cite{coco}. In the “random prompt” setting, ground truth prompts are replaced with random ones. Results in \cref{tab:coco_discuss1} show that the CLIP score for ground truth prompts (CLIP$_{\text{GT}}$) closely matches that for input prompts (CLIP$_{\text{IN}}$), indicating that SDI favors contextual coherence over strict prompt-adherence. For example, in \cref{fig:vis_random_shift:a}, SDI interprets contextual hints by generating an object riding on a motorcycle rather than a zebra as prompted, producing a person with zebra-patterned clothing instead. In the ``random prompt + shifted mask'' setting, we shift the mask by 25 pixels to add more ground-truth information into the image condition $z^c$. This adjustment decreases instruction-following accuracy, reflected by a 3\% drop in CLIP$_{\text{IN}}$ and a 5\% increase in CLIP$_{\text{GT}}$ compared to the ``random prompt'' case. In \cref{fig:vis_random_shift:b}, when an object, like a person or tie, is visible in $z^c$, SDI can revert the whole object. This analysis confirms that \textbf{the context provided by $z^c$ significantly limits SDI's instruction following}.

\begin{table}[t]
\small
\centering
\begin{tabular}{ccccc}
    \toprule
    &  AS & LPIPS & CLIP$_{\text{IN}}$ & CLIP$_{\text{GT}}$ \\
    \cmidrule(lr){1-5}
    Original & 5.89 & 0.04 & 18.66 & 18.66\\
    Shift Mask & 5.89 & 0.04 & 18.61 & 18.61\\
    \rowcolor{blue!20} %
    Random Prompt & 5.79 & 0.04 & 15.57 & 15.62 \\
    \rowcolor{blue!20} %
    Random + Shift & 5.66 & 0.03 & 15.09 & 16.38\\
    \bottomrule
\end{tabular}
\caption{Table of SDI model on different settings toward inpainting COCO dataset. CLIP$_{\text{IN}}$ denotes the CLIP similarity toward the input prompt, while CLIP$_{\text{GT}}$ for the ground truth prompt.}
\label{tab:coco_discuss1}
\vspace{-10pt}
\end{table}

\begin{figure}[t]
    \begin{subfigure}[b]{\linewidth}
         \centering         \includegraphics[width=\linewidth]{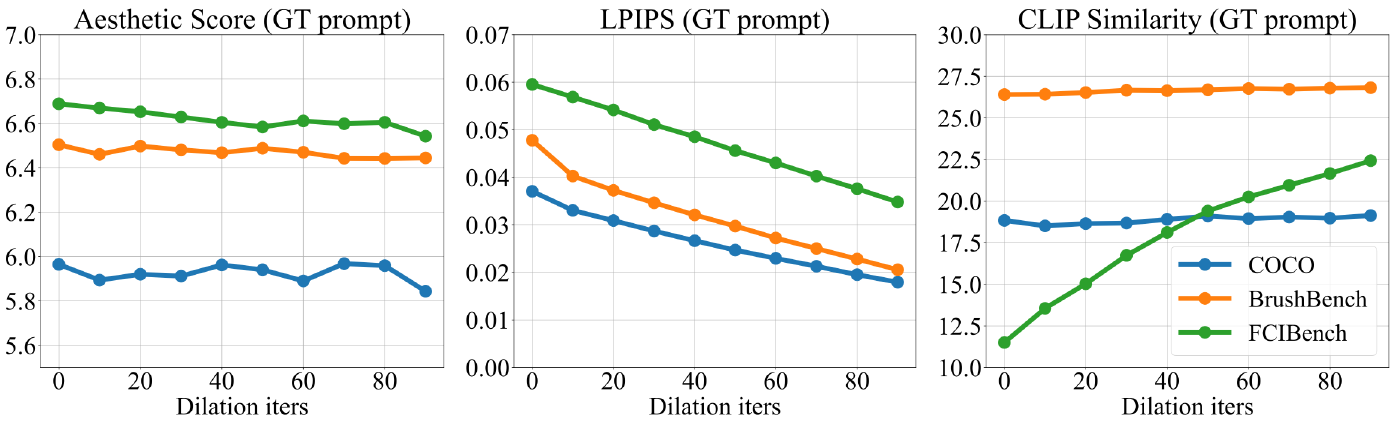}
         \caption{Ground Truth Prompt + Mask Dilation}
         \label{fig:mask_dilation:a}
    \end{subfigure}
    
    \vspace{6pt}
    
    \begin{subfigure}[b]{\linewidth}
         \centering
         \includegraphics[width=\linewidth]{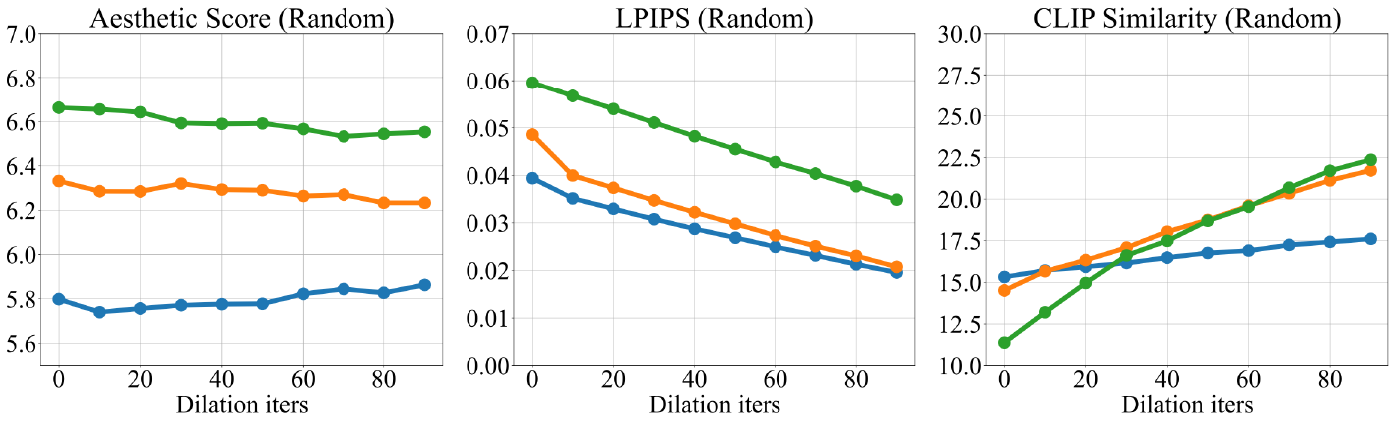}
         \caption{Random Prompt + Mask Dilation}
         \label{fig:mask_dilation:b}
        
    \end{subfigure}
    \caption{Illustration of mask size impact on inpainting metrics.}
    \label{fig:mask_dilation}
    \vspace{-10pt}
\end{figure}

\subsection{Influence of Input Mask $M$}\label{sec:d2}

In the preceding analysis, we observe that the inpainting result can be hugely guided by the image condition $z^c$, leading to its deficiency in the instruction following. Here, we explore how to adjust the input mask $M$ to promote instruction-following outputs across varied inputs. Studying the SDI model under complex scenarios—such as multiple or rough masks, unrelated prompt instruction $p$ for reference image $I$—requires a more comprehensive benchmark. However, since the COCO dataset includes only the precise masks, and its prompts for generating are simple and highly related to the image context, we propose FCIBench, which incorporates rough masks, multi-mask, and complex prompts with unrelated contexts of image condition. FCIBench compensates the shortage of existing benchmarks~\cite{brushnet,coco}, as shown in Appendix.

Building on our observation in the first row of \cref{fig:vis_random_shift:a} that the prompt \textit{``zebra''} and ground-truth \textit{``person''}  coexist, we explore which modifications to the input mask $M$ can facilitate this coexistence across different scenarios. Intuitively, we hypothesize that \textbf{increasing the mask size may provide SDI with more potential solutions to integrate prompt instructions into the image context, rather than simply disregarding the prompt}. The results, illustrated in \cref{fig:mask_dilation}, reveal that in both scenarios, AS remains nearly constant, indicating that image quality is nearly invariant to mask size. Additionally, LPIPS decrease as the non-masked area became smaller. Finally, as mask size increases, CLIP consistently improves, especially in \cref{fig:mask_dilation:b} where the prompt was unrelated to the background context. This finding supports our hypothesis that \textbf{increasing the mask size can enhance prompt-adherence in the model's output}.

\subsection{The Mechanism Behind Inpainting}
\label{sec:d3}
In \cref{sec:d1}, we identify that SDI's deficiency in instruction-following stems from its preference for maintaining harmony within the image context. Although in \cref{sec:d2} we show that simply enlarging the mask $M$ can better balance instruction-following and contextual harmony, this solution is impractical, as we seek to modify only specific regions within the given mask. This discussion raises two key questions: \textbf{(1) How does the image condition influence features within the masked area?} and \textbf{(2) How does prompt information selectively affect only the masked area?} By uncovering the underlying mechanisms behind these questions, we can explore ways to refine SDI’s learned behavior.

\begin{figure}[t]
    \centering
   \includegraphics[width=1\linewidth]{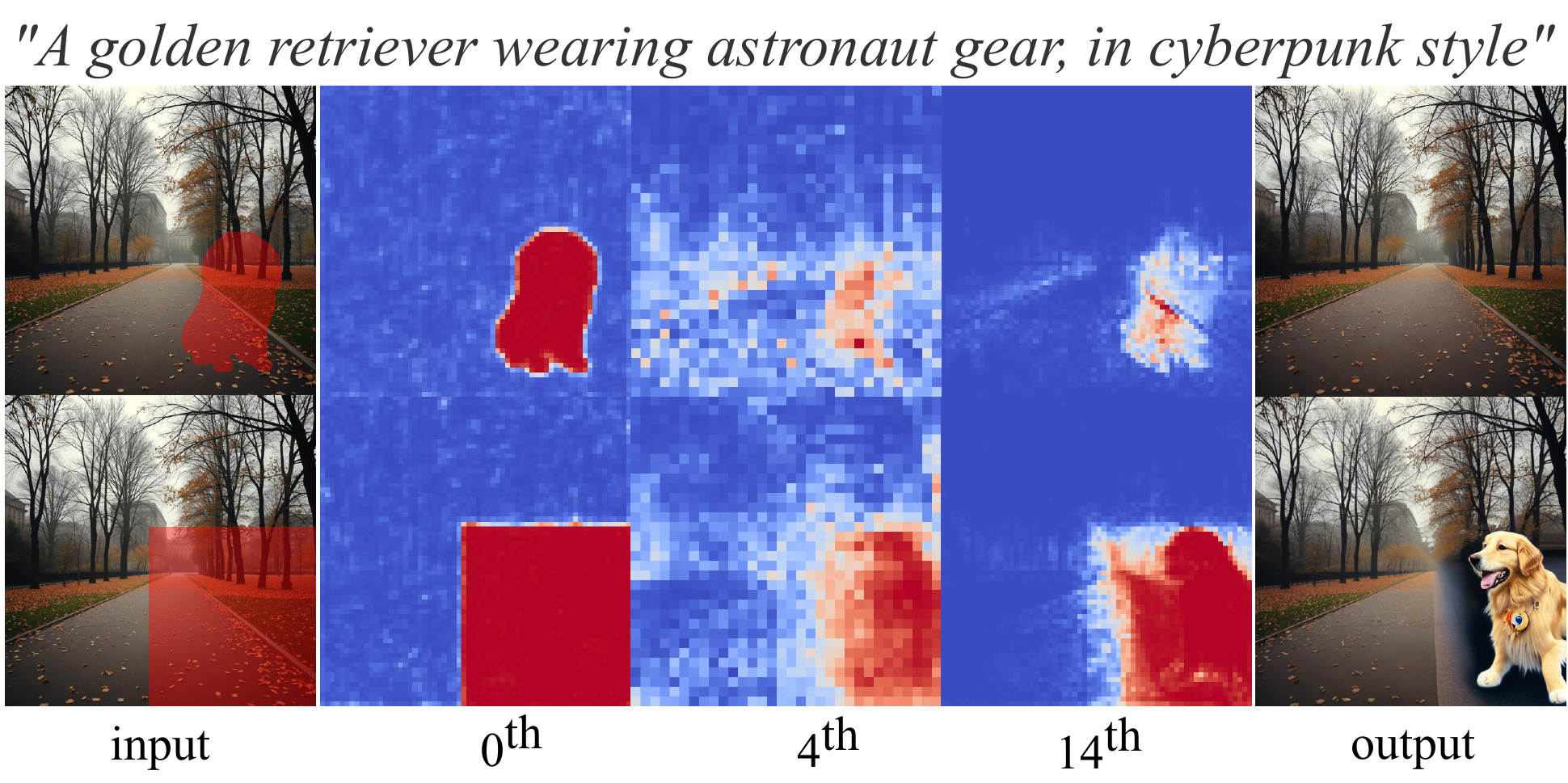}
    \caption{A self-attention visualization in different layers. The attention from $M$ is \textcolor{red}{colored}.}
    \label{fig:self_attn_vis}
    \vspace{-10pt}
\end{figure}

Our initial hypothesis to the first question is that \textbf{features within the masked area become progressively diluted by background elements during down-sampling and self-attention operations}. This is illustrated in \cref{fig:self_attn_vis}, where we compared two cases, ``precise mask'' (row 1) and ``large rough mask'' (row 2). For both cases, the mask shape is clearly visible in early layers. However, in the subsequent layers (4th, 14th) of the first case, attention within the masked area becomes further diluted by background elements. This results in the final image output with background-like elements in the masked area that are totally unrelated to the prompt. By contrast, in case 2, the attention of the generated object within the masked area successfully deviates from the background elements, aligning closely with the generated object. This results in a more instruction-following outcome, supporting our hypothesis.

\begin{figure}[t]
    \centering
    \includegraphics[width=1\linewidth]{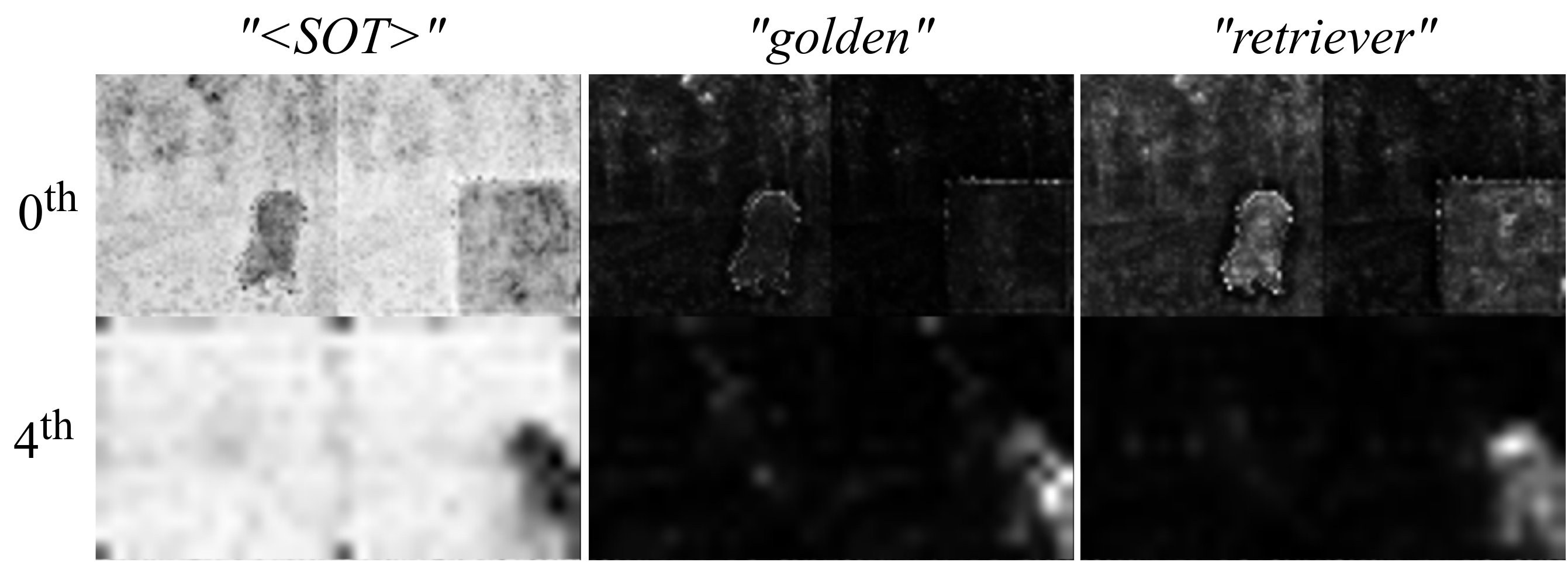}
    \caption{A cross-attention visualization of \cref{fig:self_attn_vis} in different cross-attention layers. The attention follows the input mask shape in the first layer, adapting to the output shape in the deeper layer.}
    \label{fig:cross_vis}
    \vspace{-10pt}
\end{figure}


To address the second question, we hypothesize that, given the architecture of SDI, \textbf{certain channels within the cross-attention key are highly adapted to the mask input, selectively enhance prompt response toward the masked region}. This adaptation is illustrated in \cref{fig:cross_vis} and is supported by the classifier-free guidance differences observed in Appendix. 
To strengthen this hypothesis, we measure the numerical influence made by input $M$. As we mentioned in \cref{sec:pre}, the mask input $M$ is processed into $M^c$ and $z^c$. In the initial input layer of UNet, these inputs are \text{Concat}enated and forwarded into $h_0 = \Phi_0(\text{Concat}([z_t, M^c, z^c])) \in \mathbb{R}^{(H/4) \times (W/4) \times 320}$, where $\Phi_0$ is the first convolutional layer of the denoising UNet. Following this, the cross-attention layer projects $h_0$ feature into the query representation $Q = W_Q h_0$, where $Q \in \mathbb{R}^{(H/4 \times W/4) \times 320}$. Simultaneously, the prompt embedding $p$ is projected into $K = W_K\Psi(p)$, $V = W_V\Psi(p)$, where $\Psi$ is the CLIP text encoder
and $K, V \in \mathbb{R}^{77 \times 320}$. The cross-attention output is then computed as $\text{Attention}(Q,K,V) = \text{Softmax}(QK^T/\sqrt{d})V$.


To test our assumption that specific channels of query $Q$ are highly adapted to the given key token $k \in \mathbb{R}^{320}$ in certain feature channels, we define a Channel Influence Indicator ($CI$) here: 
\begin{equation}
CI(Q, M, k, i) = \frac{1}{\sum_{j} \bar{M_j}}\sum_{j=1}^{{H \times W}/{16}} (Q_j \odot k)_i \cdot \bar{M}_j,
\end{equation}
where $\odot$ denotes the Hadamard product, the subscript $i$ refers to the $i$-th element of a vector, and $\bar{M}$ represents the flattened version of $M$. Since the sum of $CI$ across different channels is positively correlated with the $QK^T$ computation in cross-attention, the $CI$ indicator offers a means to visualize the influence introduced by $M$ within specific feature channels.

\begin{figure}[t]
  \centering
   \includegraphics[width=1\linewidth]{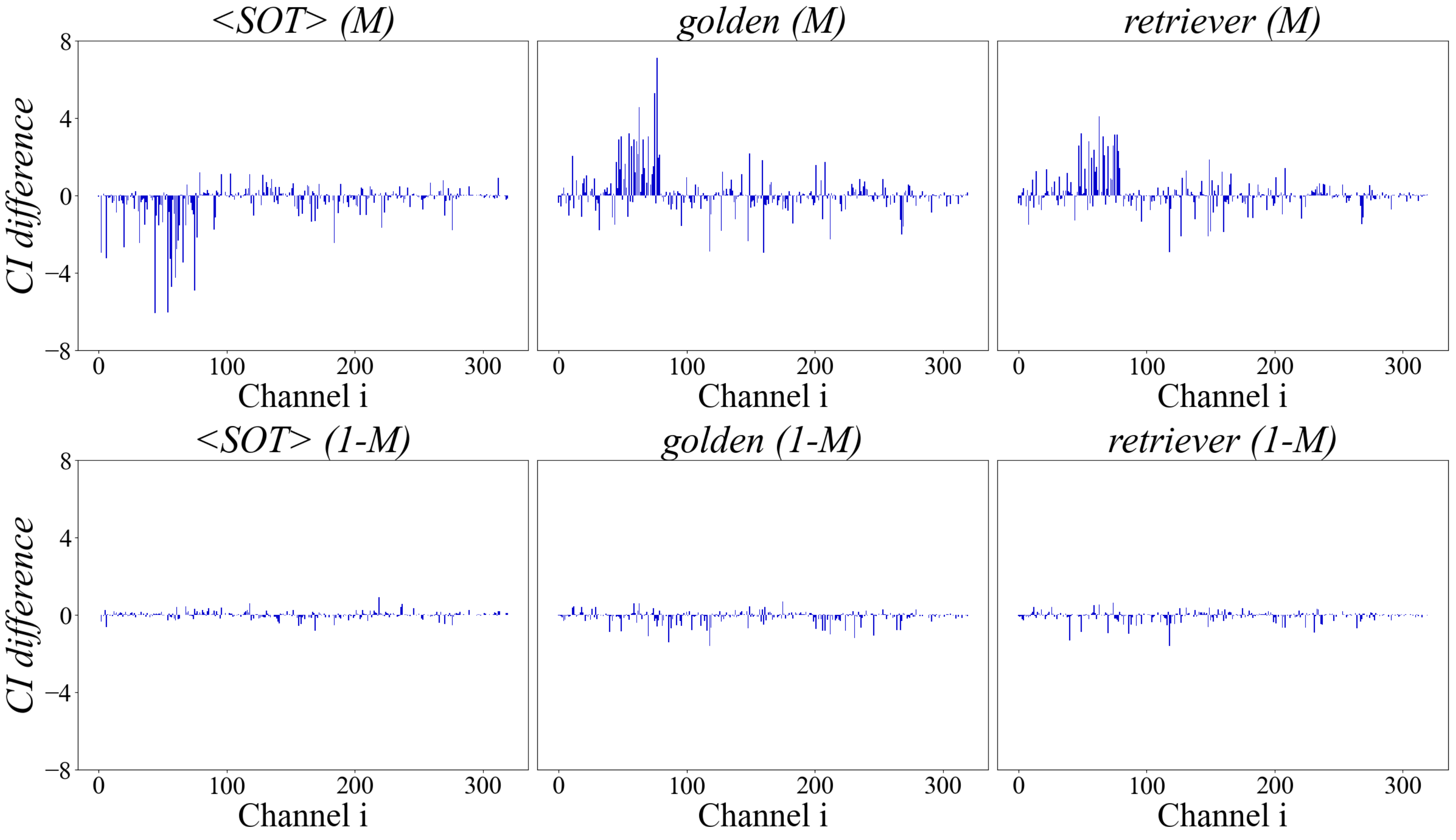}
   \caption{An illustration of the difference of channel influence indicator ($\Delta CI$) across different channels, in the region $M$ (first row) and the region $(1-M)$ (second row).}
   \label{fig:ci_plot}
   \vspace{-10pt}
\end{figure}

To measure the difference leading by mask input $M$, we choose the input mask of the second row of \cref{fig:self_attn_vis}, denoted as $M^l$, compared with the zero matrix $M^n$. We then define $Q^{\text{l}} = W_Q\Phi_0(\text{Concat}([z_t, M^{\text{l}}, z^{\text{l}}])$ and $Q^{\text{n}} = W_Q\Phi_0(\text{Concat}([z_t, M^{\text{n}}, z^{\text{n}}])$, where $z^{\text{l}}$ is the image condition given $M^{\text{l}}$, similarily the $z^{\text{n}}$ and $M^{\text{n}}$.

To demonstrate that mask input ($M^l$ in this case) significantly influences $QK^T$ computations in certain channels of given masked area $M^l$, we plot $\Delta CI = CI(Q^{\text{l}} , M^{\text{l}}, k, i) - CI(Q^{\text{n}}, M^{\text{l}}, k, i)$, at initial timestep t=T, where both $Q^{\text{l}}$ and $Q^{\text{n}}$ share the same noise latent $z_t$. In \cref{fig:ci_plot}, we observe that the $\Delta CI$ shifts more markedly within $M^{\text{l}}$ area than in $(1-M^{\text{l}})$ area. For non-informative tokens (\ie ``$<$\textit{SOT}$>$''), $\Delta CI$ decreases significantly, leading to a relative increase in attention toward other informative tokens. For meaningful tokens such as ``\textit{golden}'' and ``\textit{retriever}'', $\Delta CI$ increases, especially within the first 80 channels. This finding supports our hypothesis that \textbf{cross-attention key features adapt specifically to mask input, particularly in the first 80 channels, enabling selective prompt influence within $M$}. More evaluations can be found in Appendix.
\begin{figure}[t]
  \centering
   \includegraphics[width=1\linewidth]{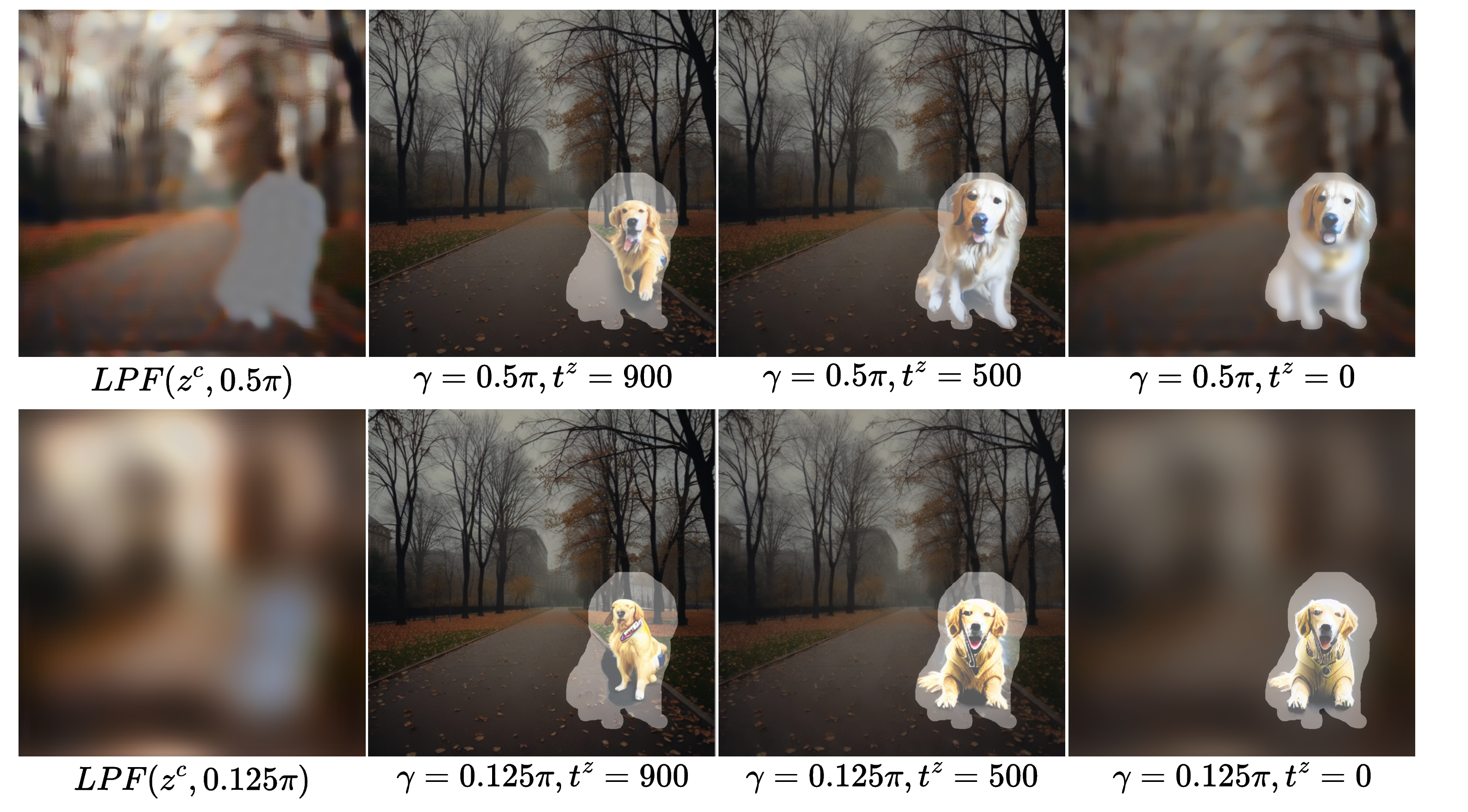}
   \caption{A illustration of the $z^{fc}$ (column 1) and the output leading by different values of $t^{z}$ (columns 2,3,and 4). The input mask is highlighted by overlaying it onto the output images.}
   \label{fig:zfc_example}
   \vspace{-10pt}
\end{figure}

\section{Method}
\label{sec:method}

In \cref{sec:d1}, we identify that the image context provided by $z^c$ can impede instruction-following in the SDI model. In \cref{sec:d2} and \cref{sec:d3}, we observe that the $M^c$ plays an important role in the cross-attention layer, the inclusion of $M^c$ leading to prompt-adherence by shifting the cross-attention features. Building on these insights, we propose FreeCond to directly reduce the heavy reliance on the image context provided by $z^c$ and increase the feature shift led by $M^c$. With FreeCond, we can achieve improved instruction-following for the SDI-based approach in a post-hoc manner without additional fine-tuning or computational costs.

\subsection{FreeCond Image Condition}
In light of the phenomenon observed in \cref{sec:d1}, where inpainting outputs can be significantly influenced or dominated by the image condition $z^c$, it appears reasonable to reduce the influence of $z^c$ to improve insturction following. However, since the inpainting model relies on $z^c$ to preserve the background, any adjustments to $z^c$ can compromise the mask preservation. Nonetheless, based on the nature of the T2I diffusion process, as described in \cite{understanding_t2i, freeu, ediff}, we note that low-frequency components are formed in early steps while high-frequency details emerge in later steps. In other words, we can still largely preserve the background in the final output by inputting only the low-frequency portion of $z^c$ in the early step and then transitioning to the original $z^c$. We define the FreeCond image condition as:
\begin{equation}
z^{fc} = 
\begin{cases} 
LPF(z^c, \gamma), & \text{if } t \geq t^{fc} \\
z^c, & \text{if } t < t^{fc} 
\end{cases}
\end{equation}
where $LPF(z^c, \gamma)$ is a low-pass filter that excludes high-frequency components above the threshold $\gamma$, and $t^{fc}$ is the timestep control parameter, with lower values of $t^{fc}$ producing a blurrier output. The effect of $z^{fc}$ is demonstrated in \cref{fig:zfc_example}. By modifying $z^{fc}$ in the early step, such as setting $t^{fc} \in [0.5T, 0.9T]$, we can effectively improve instruction-following with minimal impact on background preservation. Since the $LPF$ filters out high-frequency image information, the overall image context is disrupted, thus enhancing instruction-following by reducing interference from the original image context. 

\begin{figure}[t]
    \centering
    \begin{subfigure}[t]{\linewidth}
        \centering
        \includegraphics[width=\linewidth]{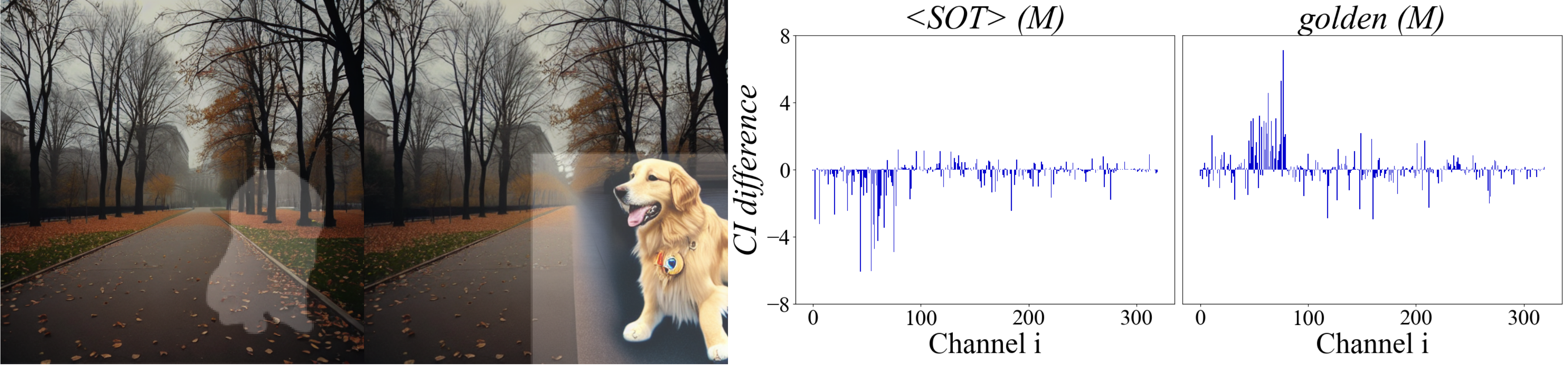}
        \caption{$\alpha=1,\beta=0$}
        \label{fig:mfc_example:a}
    \end{subfigure}
    
    \begin{subfigure}[t]{\linewidth}
        \centering
        \includegraphics[width=\linewidth]{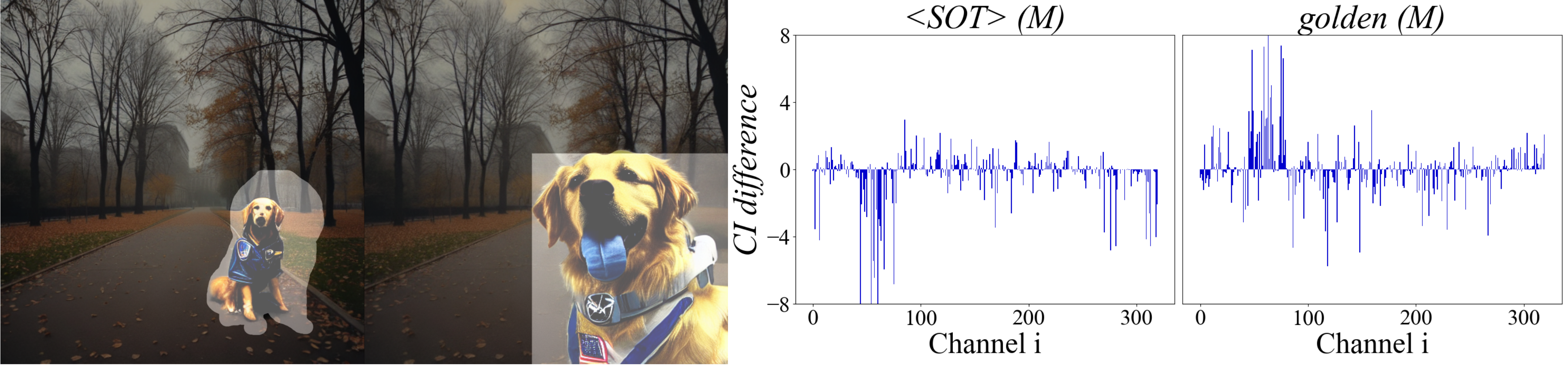}
        \caption{$\alpha=3,\beta=0$}
        \label{fig:mfc_example:b}
    \end{subfigure}
    
    \begin{subfigure}[t]{\linewidth}
        \centering
        \includegraphics[width=\linewidth]{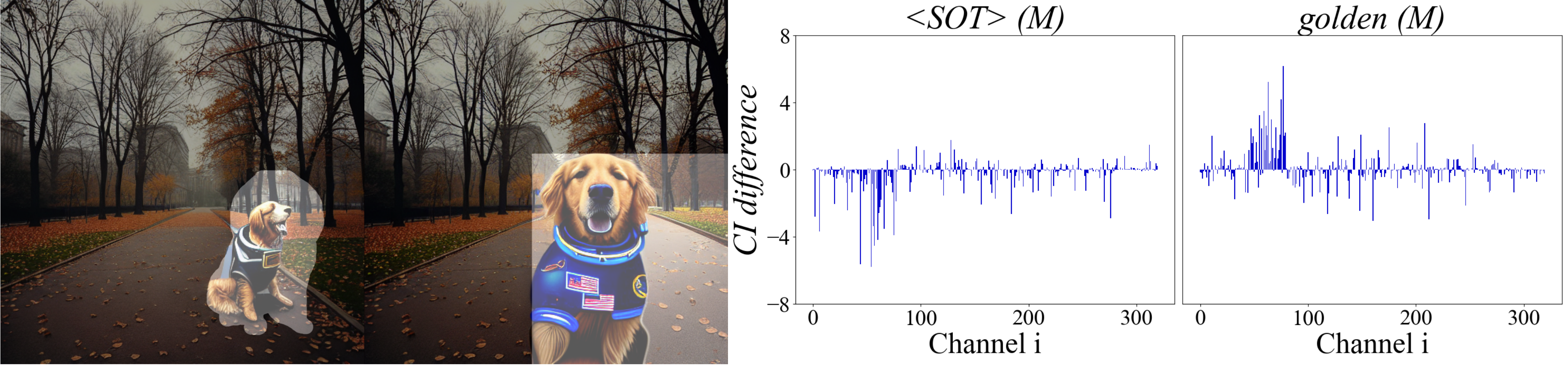}
        \caption{$\alpha=1,\beta=0.5$}
        \label{fig:mfc_example:c}
    \end{subfigure}
    \caption{A illustration of the effect of $M^{fc}$ and corresponding CI plot for ``large mask'' (columns 3 and 4). The input mask is highlighted by overlaying it onto the output images.}
    \label{fig:mfc_example}
    \vspace{-10pt}
\end{figure}

\begin{table*}[t]
    \centering
    \footnotesize
    \begin{tabular}{lccccccc}
        \toprule
        \multirow{2}{*}[-3pt]{Methods} & \multicolumn{3}{c}{Image Quality} & \multicolumn{2}{c}{Background Preservation} & \multicolumn{2}{c}{Instruction Following} \\
        \cmidrule(lr){2-4}\cmidrule(lr){5-6}\cmidrule(lr){7-8}
         & ImageReward$\uparrow$ & HPS $\uparrow$ & Aesthetic$\uparrow$ & PSNR$\uparrow$ & LPIPS $\downarrow$ & CLIP Score $\uparrow$ & IoU Score$\uparrow$ \\
        \cmidrule(lr){1-8}
        \rowcolor{blue!10} %
        SDI~\cite{stablediffusion}& -1.21/1.22/-1.95 & 0.24/0.27/0.17 & \textbf{5.90}/6.50/\textbf{6.69} & 25.95/27.26/25.54 & \textbf{0.04}/\textbf{0.04}/\textbf{0.06}  & 18.67/26.41/11.45 & 0.49/0.62/0.07\\
        \rowcolor{blue!20} %
        SDI$^{fc}$ & -1.23/1.14/-1.29 & 0.24/0.27/0.20 & 5.86/\textbf{6.54}/6.44 & 24.79/26.73/24.58 & \textbf{0.04}/0.05/0.07 & 18.82/26.46/18.27 & 0.65/0.70/0.54\\
        \cmidrule(lr){1-8}
        CNI~\cite{controlnet} & -1.24/1.15/-1.95 & 0.24/0.27/0.17 & 5.78/6.44/6.63 & \textbf{26.23}/\textbf{27.69}/\textbf{25.81} & \textbf{0.04}/\textbf{0.04}/\textbf{0.06} & 18.83/26.04/11.38 & 0.57/0.60/0.12\\
        CNI$^{fc}$ & -1.24/1.11/-1.52 & 0.24/0.27/0.19 & 5.76/6.46/6.49 & 25.58/27.06/25.23 & \textbf{0.04}/0.05/0.07  & 19.10/25.97/16.01 & 0.67/0.67/0.45\\
        \cmidrule(lr){1-8}
        HDP~\cite{hdpainter} & -1.19/1.18/-1.40 & \textbf{0.25}/0.27/0.20 & 5.79/6.46/6.57 & 24.95/27.02/25.23 & \textbf{0.04}/0.05/\textbf{0.06}  & 19.14/26.45/17.08& 0.60/0.63/0.36\\
        HDP$^{fc}$ & -1.27/1.20/-1.15 & 0.24/\textbf{0.28}/0.20 & 5.75/6.52/6.43 & 23.63/26.00/23.48 & 0.05/0.05/0.08 & 19.17/26.41/19.37& 0.77/0.68/0.67\\
        \cmidrule(lr){1-8}
        \rowcolor{blue!10} %
        PP~\cite{powerpaint} & -1.19/1.21/-1.13 & 0.22/0.27/0.19 & 5.79/6.30/6.40 & 25.63/27.62/25.29 & 0.05/0.05/0.07 & 18.74/27.02/19.05& 0.59/0.55/0.43\\
        \rowcolor{blue!20} %
        PP$^{fc}$ & \textbf{-1.15}/1.20/-1.12 & 0.22/0.27/0.19 & 5.73/6.33/6.39 & 25.41/27.37/24.32 & 0.05/0.05/0.08 & 19.12/\textbf{27.05}/19.43& 0.67/0.59/0.52\\
        \cmidrule(lr){1-8}
        \rowcolor{blue!10} %
        BN~\cite{brushnet} & -1.24/\textbf{1.24}/-1.08 & 0.24/0.27/\textbf{0.21} & 5.77/6.53/6.38 & 24.89/26.37/24.35 & 0.05/0.06/0.07 & 19.22/26.50/19.96 & 0.83/\textbf{0.75}/0.77\\
        \rowcolor{blue!20} %
        BN$^{fc}$ & -1.31/1.21/\textbf{-1.05} & 0.23/0.27/\textbf{0.21} & 5.77/6.53/6.43 & 24.21/25.38/23.49 & 0.05/0.06/0.08  & \textbf{19.27}/26.50/\textbf{20.18}& \textbf{0.85}/\textbf{0.75}/\textbf{0.78}\\
        \cmidrule(lr){1-8}\morecmidrules\cmidrule(lr){1-8}
        SDXL~\cite{sdxl} & -1.06/1.32/-1.72 & 0.25/0.29/0.19 & 5.74/6.40/6.55 & 24.61/25.78/25.00 & 0.03/0.03/0.04 & 19.09/26.96/14.16& 0.53/0.68/0.09\\
        SDXL$^{fc}$ & -0.94/1.30/-0.78 & 0.25/0.29/0.22 & 5.69/6.34/6.56 & 24.15/26.12/24.59 & 0.04/0.04/0.05 & 19.77/27.16/22.36& 0.60/0.64/0.44\\
        \toprule
    \end{tabular}
    \caption{Quantitative results showing improvements achieved by FreeCond (denoted with $^{fc}$) across three benchmarks: \textbf{COCO, BrushBench, and FCIBench, separated by "/" respectively}. Note: we discovered that the BrushBench results reported in \cite{brushnet} were calculated with an NSFW detector; as NSFW detection may vary, we disable it here to ensure a more precise evaluation.}
    \label{tab:main_table}
    \vspace{-10pt}
\end{table*}

\subsection{FreeCond Mask Condition}
Building on our observations in \cref{sec:d3}, we find that the T2I inpainting effect of the SDI model is raised by the shifting in cross-attention features with non-zero mask input $M$. Further analysis in Appendix reveals that while both $z^c$ and $M^c$ affect the inpainting outcome, shifts in cross-attention features are primarily driven by $M^c$ values. Based on this, we explore the potential to enhance cross-attention feature shifts by manipulating the mask condition $M^c$. Accordingly, we introduce a FreeCond mask condition, $M^{fc}$, which scales up the value of $M^c$ to induce a stronger cross-attention feature shift, thereby improving prompt alignment. Another observation, shown in \cref{fig:ci_plot}, is that the CI indicator in the $(1-M)$ region is subtly impacted by the mask $M$. Thus, increasing mask values within the $(1-M)$ region can amplify feature shifts within $M$. To test this, we define the FreeCond mask condition:
\begin{equation}
 M^{fc}  = \alpha \cdot M^c + \beta \cdot (1-M^c)   
\end{equation}
where $\alpha$ and $\beta$ are scaling factors to control the influence of $M^{c}$ and $(1-M^{c})$. The impact of $M^{fc}$ is illustrated in \cref{fig:mfc_example}. Compared to the baseline results in \cref{fig:mfc_example:a}, the output with a $M^{fc}$ exhibits greater prompt-adherence. For instance, in the ``precise mask'' condition, instead of filling the background element, the \textit{``golden retriever wearing astronaut gear''} appears. Additionally, in the ``large mask'' setting, the golden retriever now includes the \textit{``astronaut gear''}. We also provide the response of the CI indicator for $M^{fc}$ in the right side of \cref{fig:mfc_example} to show that modifying the latent mask $M^c$ can effectively enhance feature shifts within the cross-attention layer.

With our proposed alteration, the noise prediction function from \cref{eq:cfg} can be generalized as $\hat{\epsilon}_{{\theta}}(z_t, z^{fc},M^{fc},t,p)$. As FreeCond only modifies the input, it is compatible with other SDI-based models, detailed in Appendix.

\hspace{3cm}
\begin{figure*}[]
    \centering
    \begin{subfigure}[t]{0.0184\linewidth}
        \centering
        \includegraphics[width=1.0\linewidth]{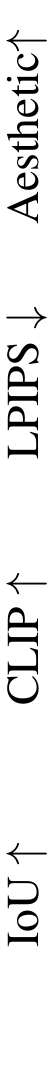}
    \end{subfigure}
    \begin{subfigure}[t]{0.183\linewidth}
        \centering
        \includegraphics[width=1.0\linewidth]{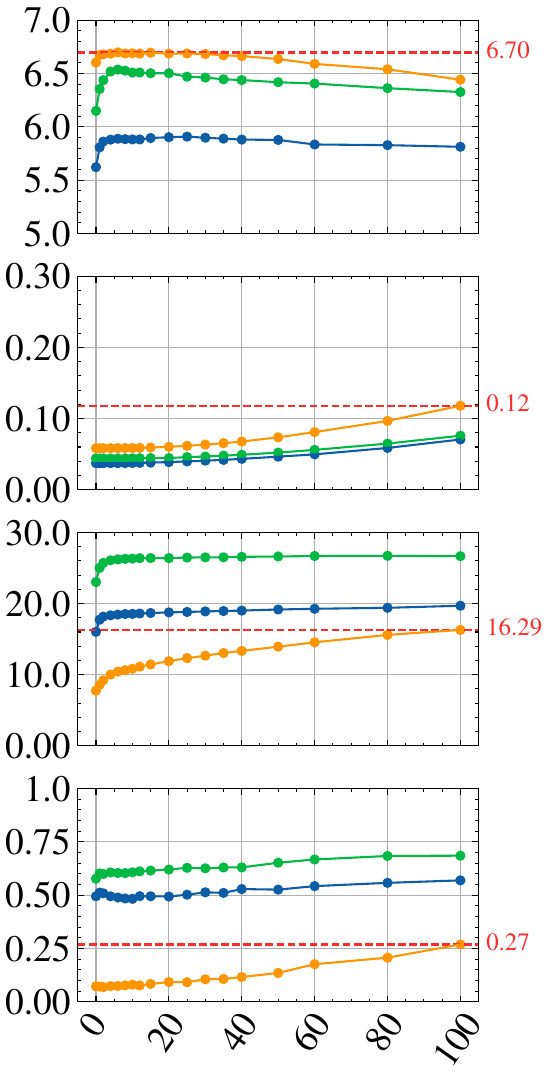}
        \caption{Change $w$.}
        \label{fig:exp_sensitive:a}
    \end{subfigure}
    \begin{subfigure}[t]{0.161\linewidth}
        \centering
        \includegraphics[width=1.0\linewidth]{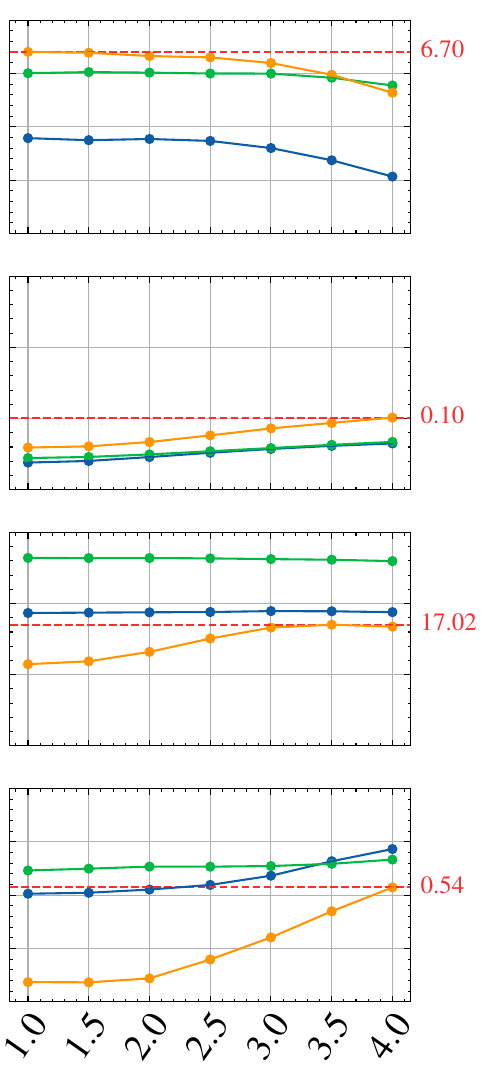}
        \caption{Change $\alpha$.}
        \label{fig:exp_sensitive:b}
    \end{subfigure}
    \begin{subfigure}[t]{0.1595\linewidth}
        \centering
        \includegraphics[width=1.0\linewidth]{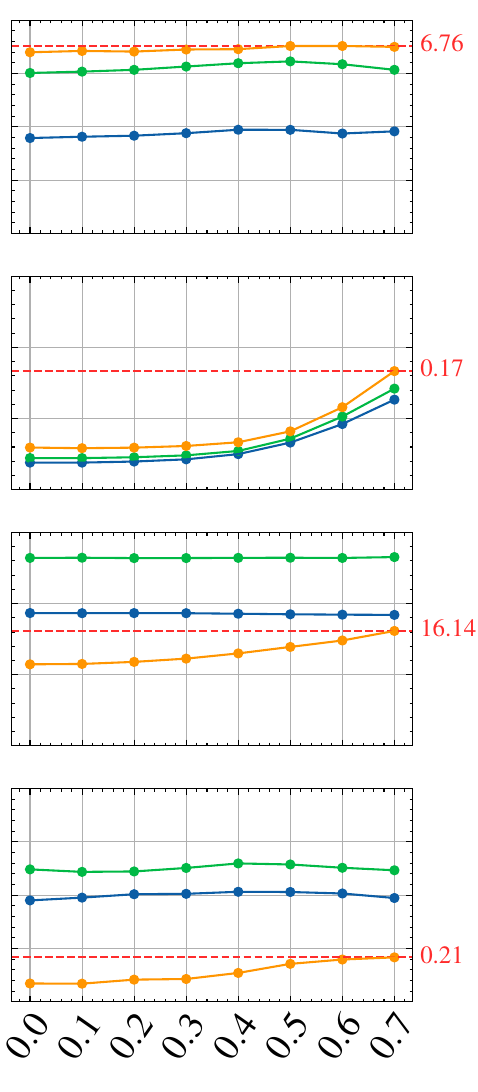}
        \caption{Change $\beta$.}
        \label{fig:exp_sensitive:c}
    \end{subfigure}
    \begin{subfigure}[t]{0.161\linewidth}
        \centering
        \includegraphics[width=1.0\linewidth]{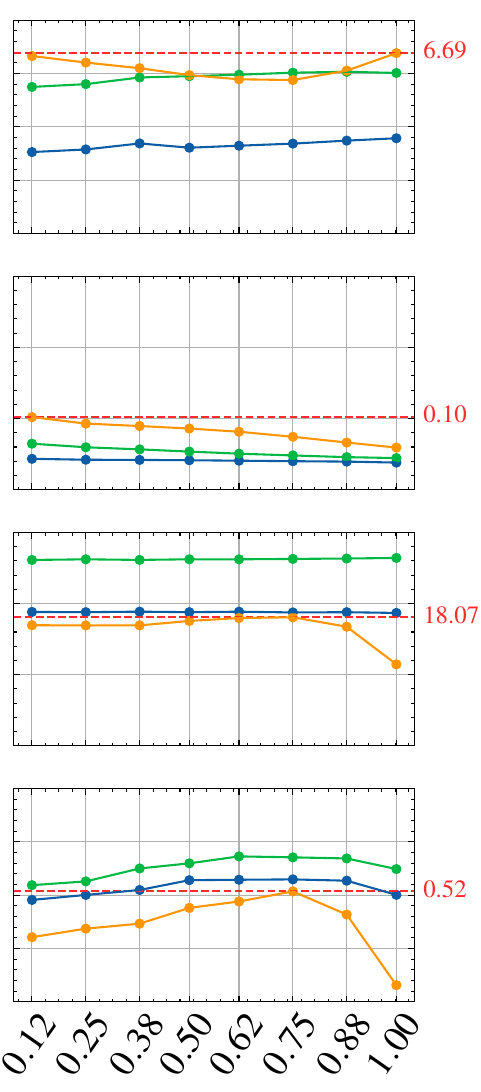}
        \caption{Change $\gamma$.}
        \label{fig:exp_sensitive:d}
    \end{subfigure}
    \begin{subfigure}[t]{0.161\linewidth}
        \centering
        \includegraphics[width=1.0\linewidth]{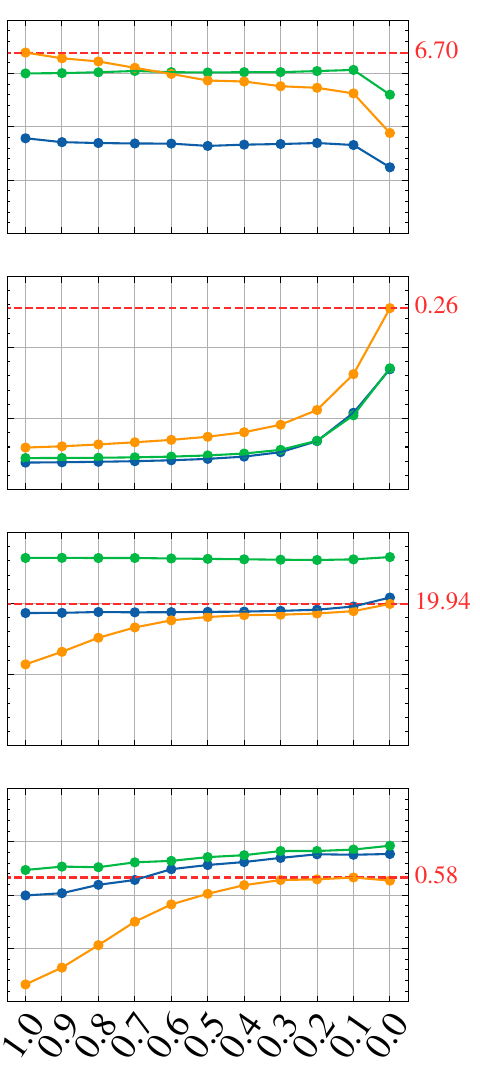}
        \caption{Change $t^{fc}$.}
        \label{fig:exp_sensitive:e}
    \end{subfigure}
    \begin{subfigure}[t]{0.07\linewidth}
        \centering
        \includegraphics[width=1.0\linewidth]{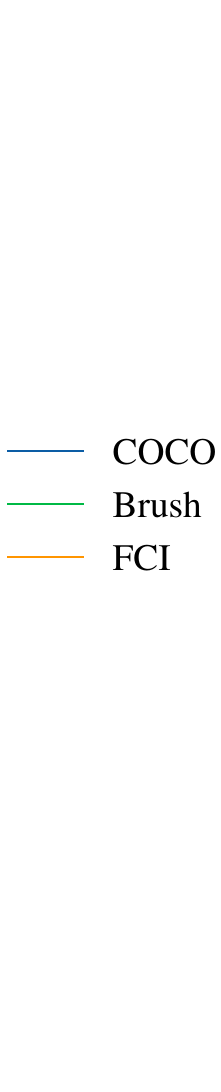}
    \end{subfigure}
    \caption{Illustration of the influence of CFG ($w$) and each hyperparameter of FreeCond($\alpha, \beta, \gamma, t^{fc}$), highest values are \textcolor{red}{denoted}.}
    \label{fig:exp_sensitive}
\end{figure*}

\begin{figure*}[]
    \centering
    \begin{subfigure}[t]{0.135\linewidth}
        \centering
        \text{\scriptsize Input}
        \includegraphics[width=1.0\linewidth]{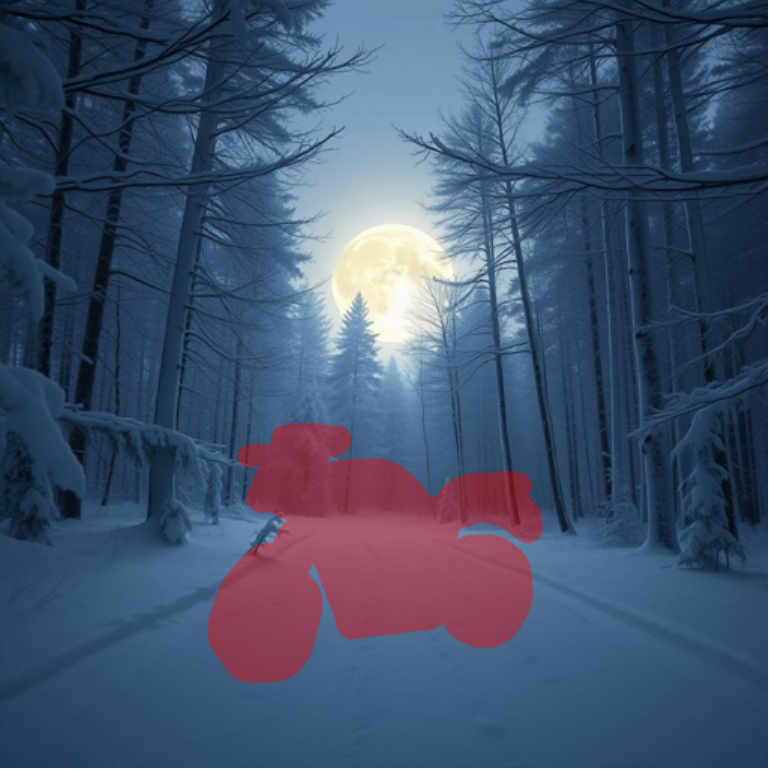}
    \end{subfigure}
    \begin{subfigure}[t]{0.135\linewidth}
        \centering
        \text{\scriptsize (15, 1, 0, $\pi$, 0)(default)}
        \includegraphics[width=1.0\linewidth]{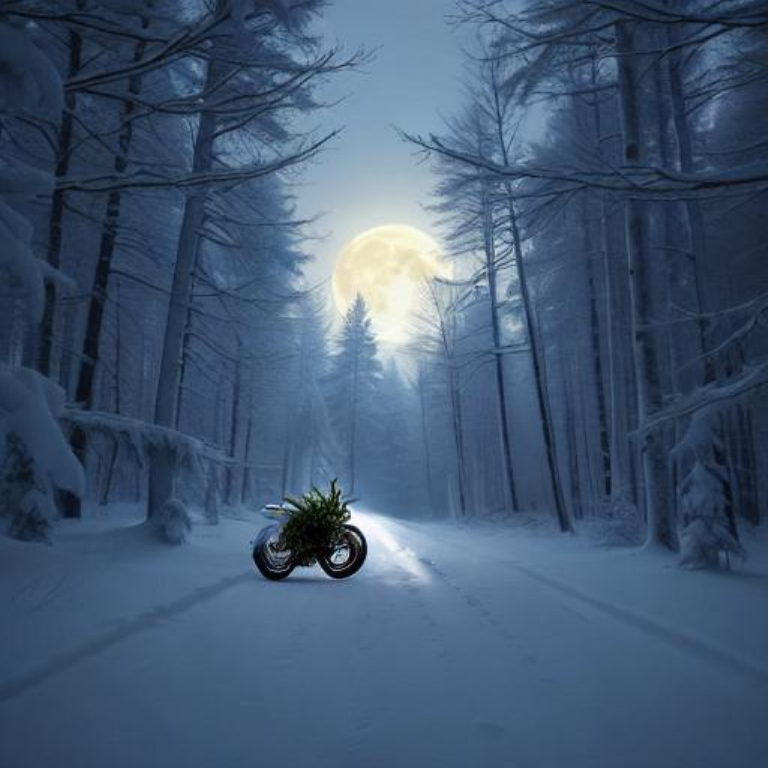}
    \end{subfigure}
    \begin{subfigure}[t]{0.135\linewidth}
        \centering
        \text{\scriptsize (\textcolor{red}{100}, 1, 0, $\pi$, 0)}
        \includegraphics[width=1.0\linewidth]{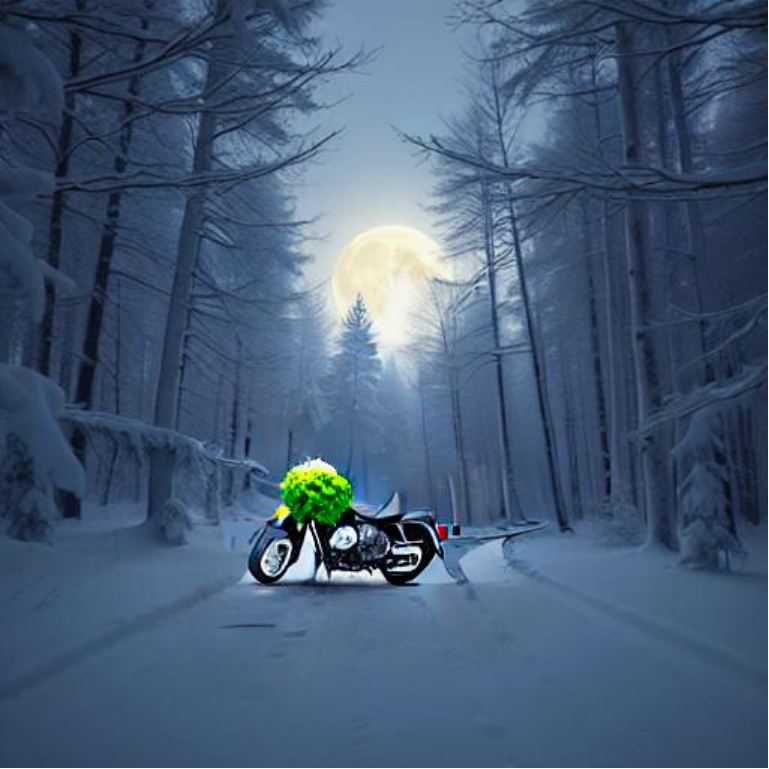}
        \caption{}
        \label{fig:exp_vis_sensitive:a}
    \end{subfigure}
    \begin{subfigure}[t]{0.135\linewidth}
        \centering
        \text{\scriptsize (15, \textcolor{red}{4}, 0, $\pi$, 0)}
        \includegraphics[width=1.0\linewidth]{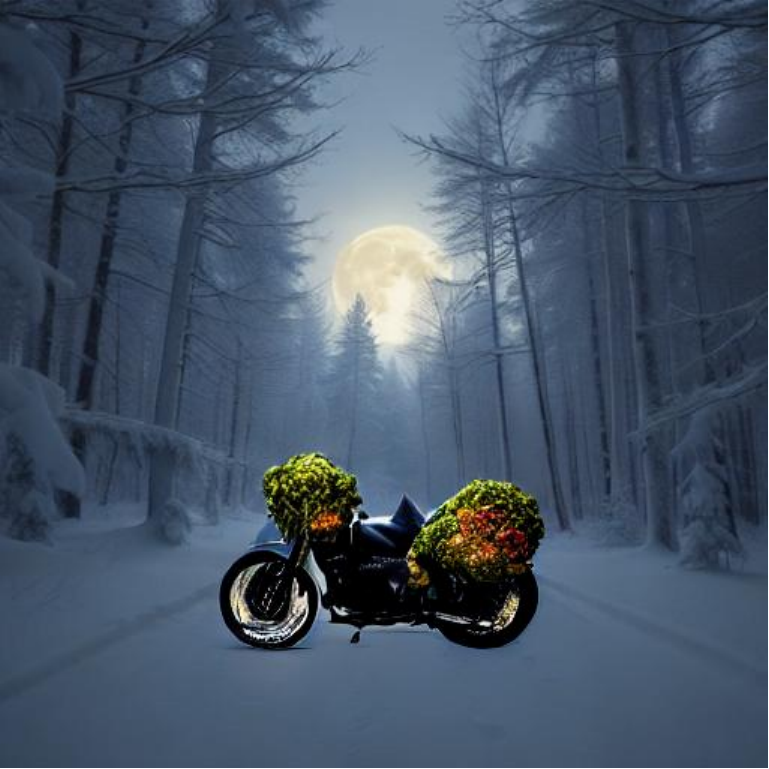}
        \caption{}
        \label{fig:exp_vis_sensitive:b}
    \end{subfigure}
    \begin{subfigure}[t]{0.135\linewidth}
        \centering
        \text{\scriptsize (15, 1, \textcolor{red}{0.7}, $\pi$, 0)}
        \includegraphics[width=1.0\linewidth]{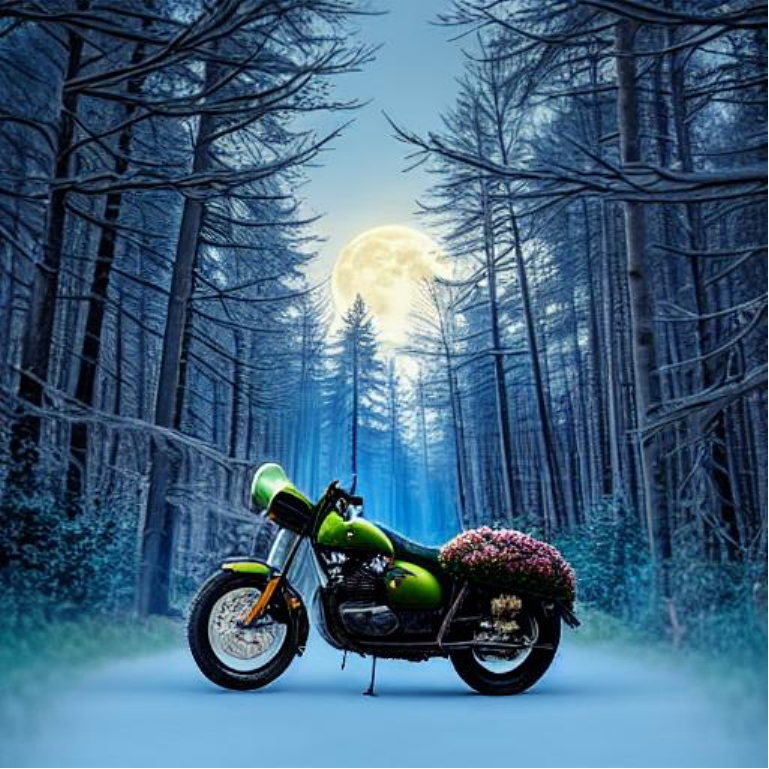}
        \caption{}
        \label{fig:exp_vis_sensitive:c}
    \end{subfigure}
    \begin{subfigure}[t]{0.135\linewidth}
        \centering
        \text{\scriptsize (15, 1, 0, \textcolor{red}{$0.75\pi$, 0.5})}
        \includegraphics[width=1.0\linewidth]{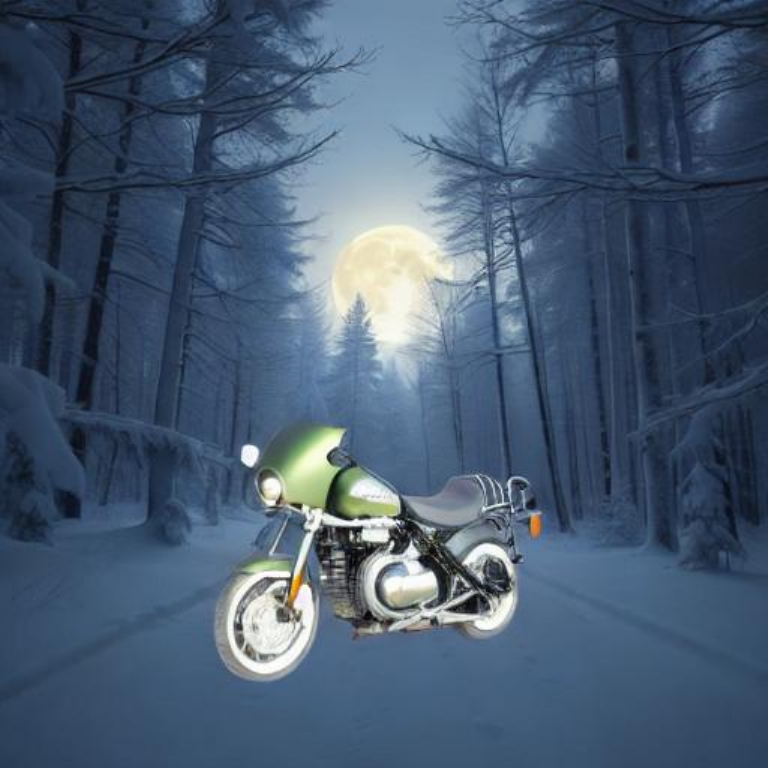}
        \caption{}
        \label{fig:exp_vis_sensitive:d}
    \end{subfigure}
    \begin{subfigure}[t]{0.135\linewidth}
        \centering
        \text{\scriptsize (15, \textcolor{red}{2, 0.35, $0.75\pi$, 0.5})}
        \includegraphics[width=1.0\linewidth]{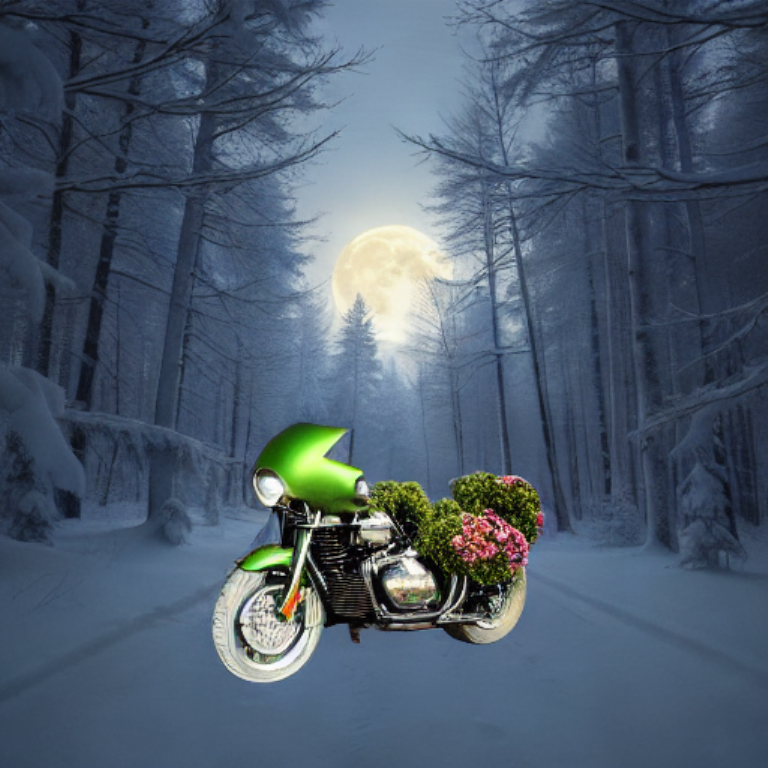}
        \caption{}
        \label{fig:exp_vis_sensitive:e}
    \end{subfigure}
    \caption{The qualitative illustration of \cref{fig:exp_sensitive}, the change compared to normal SDI is \textcolor{red}{colored}.}
    \label{fig:exp_vis_sensitive}
    \vspace{-7pt}
\end{figure*}

\section{Experiment}

\label{sec:experiment}
\subsection{Experiment Setting}
We conduct experiments on three datasets: COCO~\cite{coco}, BrushBench~\cite{brushnet}, and our proposed FCIBench, each comprising 600 instruction pairs (details in Appendix). To account for the inherent differences between the baseline methods, we provide the specific hyperparameter settings and further discussion in Appendix. The computational metrics used for evaluation are detailed in \cref{sec:pre}.

\subsection{Experiment Results} 
\cref{tab:main_table} presents the quantitative improvements achieved with the inclusion of FreeCond. We compare FreeCond with the original SDI~\cite{stablediffusion} and its variants, including ControlNet Inpainting (CNI)~\cite{controlnet}, HD-Painter (HDP)~\cite{hdpainter}, PowerPaint (PP)~\cite{powerpaint}, and BrushNet (BN)~\cite{brushnet}. Additionally, we assess SDXL~\cite{sdxl}, a much larger model, as a reference to showcase the zero-shot improvements by FreeCond, it is not directly compared with other baselines.

In our proposed FCIBench, shown in the right section of each block. Designed as a more challenging benchmark, FreeCond demonstrates substantial gains, achieving a 60\% increase over the original SDI model and a 1\% improvement over BrushNet, the existing SOTA. Additionally, FreeCond improves metrics such as IR, HPS, and IoU across all baselines. Notably, both BrushNet and PowerPaint benefit from FreeCond with modest increases in IoU yet larger gains in CLIP score, highlighting FreeCond's capability to further enhance prompt-adherence in SOTA methods that are optimized for mask-fitting. For the two widely used datasets, COCO and BrushBench—represented in the left and middle sections of each block—current inpainting methods demonstrate similar performance levels. FreeCond advances this upper limit, increasing BrushNet's CLIP score from 19.22 to 19.27 on COCO and PowerPaint's CLIP score from 27.02 to 27.05 on BrushBench. Overall, FreeCond enhances performance across both instruction following and image quality, with improvements in IR, HPS, and CLIP scores.

Nevertheless, modifying learned mechanism with FreeCond can lead to minor degradations in detail-oriented metrics, such as AS, PSNR, and LPIPS. These minor distortions, are generally imperceptible to human preference, as evidenced by the increases in the human-preference-based metrics IR and HPS. These results are more apparent in the qualitative results, discussed further in Appendix.

\subsection{Ablation Study}
\label{sec:ablation}

In \cref{fig:exp_sensitive} and \cref{fig:exp_vis_sensitive}, we examine the impact of adjusting five components: (a) the classifier-free guidance (CFG) scale $w$~\cite{cfg}, (b) the inner-mask scale $\alpha$, (c) the outer-mask scale $\beta$, (d) the LPF threshold $\gamma$ with a fixed $t^{fc}=25$, and (e) the LPF timestep $t^{fc}$ with $\gamma = 0.75\pi$. For each test, we fix the parameters at $(w, \alpha, \beta, \gamma, t^{fc}) = (15, 1, 0, \pi, T)$ (the default configuration of original SDI) and vary only one parameter at a time. Based on quantitative and qualitative outcomes, we summarize our findings below.

\noindent\textbf{Effect of $w$.} As discussed in \cref{sec:d1}, SDI’s random masking strategy prioritizes generating objects within the mask rather than strict mask conformity. Therefore, increasing $w$ in \cref{fig:exp_vis_sensitive:a} primarily enhances prompt-related details without substantially increasing object size. This outcome is further reflected in the \textbf{lesser improvement of the IoU score compared to the CLIP score} in \cref{fig:exp_sensitive:a}. 

\noindent\textbf{Effect of $\alpha$.} As explained in \cref{sec:method}, increasing $\alpha$ intensifies the cross-attention response within the masked area, \textbf{enhancing both prompt-adherence and mask-fitting}, as illustrated in \cref{fig:exp_vis_sensitive:b} and \cref{fig:mfc_example:b}. However, excessively high $\alpha$ disrupts the learned feature distribution, leading to over-saturated results and a drop in AS.

\noindent\textbf{Effect of $\beta$.} Unlike other parameters, increasing $\beta$ \textbf{enhances both the CLIP score and AS}, indicating a stronger self-attention interaction between $M$ and $1-M$, which results in a more harmonious output. However, as shown in \cref{fig:exp_vis_sensitive:c} and \cref{fig:mfc_example:c}, higher $\beta$ values also increase background distortion, reflected by the LPIPS in \cref{fig:exp_sensitive:c}.

\noindent\textbf{Effect of $z^{fc}$ (controlled by $\gamma$ and $t^{fc}$).} These parameters control the frequency components of $z^{fc}$, which play a key role in reducing contextual influence and establishing prompt-related structures at early timesteps. As shown in \cref{fig:exp_sensitive:e}, increasing $t^{fc}$ significantly \textbf{improves both CLIP and IoU scores}. This is reflected in \cref{fig:exp_vis_sensitive:d}, where mask-fitting is improved while prompt-adherence is lacking (\eg, the \textit{``moss and flowers''} are not fully generated).

\section{Conclusion}
In this work, we identify an instruction-following deficiency that persists across current SDI-based inpainting methods, particularly when complex prompts are provided alongside unrelated image conditions. Through an in-depth investigation of the SDI mechanism, we discover that its selective inpainting capability within masked areas stems from a feature shift in the cross-attention layer. Based on this insight, we propose FreeCond—a training-free plug-in that introduces no additional computation overhead. Unlike classifier-free guidance, FreeCond enhances not only prompt-adherence but also mask-fitting and image quality. However, excessive parameter adjustments can degrade image quality, highlighting the need for careful tuning.

{
    \small
    \bibliographystyle{ieeenat_fullname}
    \bibliography{main}
}

\end{document}